\documentclass[iicol]{sn-jnl-arxiv}% Math and Physical Sciences Reference Style
\jyear{2022}%
\usepackage[english]{babel}
\usepackage{cite,wrapfig}
\usepackage{amsfonts}
\usepackage{graphicx, epstopdf}
\usepackage{graphics}
\usepackage{amsmath}
\usepackage{caption}
\usepackage{bbm}
\usepackage{multirow}
\usepackage{subfig}
\usepackage{hyperref}
\usepackage{xcolor}
\usepackage{array}
\usepackage{verbatim}
\usepackage{float}
\usepackage{placeins}
\usepackage{stackengine}

\usepackage{mathtools}

\renewcommand\thefigure{\arabic{figure}}

\def\N{\mathbb{N}}

\def\R{\mathbb{R}}

\def\1{\mathbbm{1}}

\def\0{\mathbf{0}}

\def\D{\mathcal{D}}

\def\W{\mathbb{W}}

\DeclareMathOperator{\diam}{diam}

\DeclareMathOperator*{\argmax}{argmax}

\theoremstyle{thmstyleone}%
\newtheorem{theorem}{Theorem}
\newtheorem{lemma}{Lemma}%  meant for continuous numbers
\newtheorem{proposition}[theorem]{Proposition}% 

\theoremstyle{thmstyletwo}%

\theoremstyle{thmstylethree}%
\newtheorem{definition}{Definition}%

\begin{document}
\title[RPDG]{A Random Persistence Diagram Generator}

%Farzana Nasrin\thanks{Department of Mathematics, University of Hawaii at Manoa}

 %Theodore Papamarkou \footnotemark[3] \thanks{Department of Mathematics, The University of Manchester}

 %Na Gong\thanks{Department of Materials Science and Engineering, University of Tennessee, Knoxville}

 %Orlando Rios\footnotemark[3] 

 %Vasileios Maroulas\thanks{Department of Mathematics, University of Tennessee, Knoxville}

\author[3,1]{\fnm{Theodore} \sur{Papamarkou}}

%\email{t.papamarkou@manchester.ac.uk}

\author[2]{\fnm{Farzana} \sur{Nasrin}}

%\email{fnasrin@hawaii.edu}

\author[1]{\fnm{Austin} \sur{Lawson}}

%\email{alawso50@utk.edu}


\author[4]{\fnm{Na} \sur{Gong}}

%\email{ngong@utk.edu}
\author[4]{\fnm{Orlando} \sur{Rios}}
%\email{orios1@utk.edu}
\author*[1]{\fnm{Vasileios} \sur{Maroulas}}\email{vmaroula@utk.edu}

\affil[1]{\orgdiv{Department of Mathematics}, \orgname{University of Tennessee}, \orgaddress{\city{Knoxville}, \state{Tennessee}, \country{US}}}

\affil[2]{\orgdiv{Department of Mathematics}, \orgname{University of Hawai'i}, \orgaddress{\city{M\=anoa}, \state{Hawai'i}, \country{US}}}

\affil[3]{\orgdiv{Department of Mathematics}, \orgname{The University of Manchester}, \orgaddress{\city{Manchester}, \country{UK}}}
\affil[4]{\orgdiv{Department of Material Science and Engineering}, \orgname{University of Tennessee}, \orgaddress{\city{Knoxville}, \state{Tennessee}, \country{US}}}

\abstract{Topological data analysis (TDA) studies the shape patterns of data. Persistent 
homology is a widely used method in TDA that summarizes homological 
features of data at multiple scales and stores them in persistence diagrams 
(PDs). 
% As TDA is commonly used in the analysis of high dimensional data sets, a 
% sufficiently large amount of PDs that allow performing statistical analysis is 
% typically unavailable or requires inordinate computational resources.
In this 
paper, we propose a random persistence diagram generator (RPDG) method that  
generates a sequence of random PDs from the ones produced by the data. RPDG is 
underpinned  by a model 
based on pairwise interacting point processes,
and a reversible jump Markov chain Monte Carlo (RJ-MCMC) algorithm.
% Simulation of persistence diagrams - creating persistence diagrams by random 
%sampling  from the one created by the data - is an invaluable tool for many 
%statistical inferences. However, there appears to be no general distributional 
%model for persistence diagrams that can capture the specific spatial pattern 
%of 
%them as well as is capable of accommodating a sampling method to generate a 
%sequence of persistence diagrams. 
% The model combines
% a Dirichlet tesselation to capture spatial inhomogeneity of the location of points in a PD  and a step function
% to capture the
% pairwise interaction between them. 
% % For generating samples of PDs, characterized as spatial point processes, 
% % a reversible jump Markov chain Monte Carlo (RJ-MCMC) algorithm
% % is developed.
% The RJ-MCMC algorithm incorporates
% trans-dimensional 
% addition and removal of points
% and same-dimensional
% % moves based on
% relocation of points
% across samples of PDs.
A first example, which is based on a synthetic dataset,
demonstrates the efficacy of RPDG
and provides a comparison with another method for sampling PDs.
A second example demonstrates the utility of RPDG
to solve a materials science problem
given a real dataset of small sample size.
% in particular, RPDG enables hypothesis testing
% to study the processing-structure-property relationship
% of austenitic stainless steels. }
}




\maketitle

\section{Introduction} 
\label{sec:intro}

Several modern machine learning models rely on being trained on a large number of data.
However, the amount of available data is limited in many  applications,
or data generation (from experimental facilities) can be expensive or time consuming.
For example, quantitative microstructure analysis relies on data to
understand and enhance the structural properties of high strength steel;
% Several experiments are used to generate datasets required for quantitative microstructure analysis. 
the generation of these data can be very costly and time-intensive depending on the material itself or other experimental factors, such as pre-treatment of the material and test equipment. In this work, we develop a novel sampling method 
for random persistence diagram generation (RPDG)
that augments topological summaries of the data,
thus facilitating statistical analysis with limited amount of data.
We present the applicability of RPDG to a materials science problem of analyzing quantitatively
the microstructure of austenitic  stainless  steels  (AuSS) given a dataset of small sample size. 
Although we apply RPDG to analyze AuSS structured materials,
RPDG is a general method that can be employed in other applications.

% To bypass this we implement a novel sampling method RPDG that augments the data space with a sequence of persistence diagrams and allows us to perform statistical analysis such as hypothesis testing. 

Persistent homology (PH) is a  topological data analysis (TDA)
tool that provides a robust way to probe 
information about the shape of datasets and to summarize  salient features into 
persistence diagrams (PDs).
These diagrams are multisets of points in the plane, where each point represents a 
homological feature whose `time' of appearance and disappearance is contained in 
the coordinates of that point \cite{Edelsbrunner2010}. Intuitively, the 
homological features represented in a PD measure the connectedness and the void space of data as 
their resolution changes.
PH has proven to be promising in a variety of applications 
such as shape analysis \cite{Patrangenaru2018}, image analysis 
\cite{Guo2018,Love2021}, neuroscience 
\cite{Biscio2019, Nasrin2019, maroulas2019},   dynamical systems 
\cite{Khasawneh2016}, signal analysis \cite{Marchese2018}, 
chemistry and material science \cite{Maroulas2019a, Townsend2020}, and 
genetics \cite{Humphreys2019}.

There have been a number of notable contributions to develop statistical 
methods for performing inference on  topological summaries. Many of these 
methods introduce probability measures for PDs to capture statistical 
information such as means, variance and conditional probabilities 
\cite{Mileyko2011, munch2015, Turner2014}. 
Kernel densities are used by 
\cite{Bobrowski2014} to estimate PDs generated by point 
process samples drawn from a distribution. The study in
\cite{maroulas2019} 
constructs a kernel density estimator based on finite set 
statistics for nonparametric estimation of PD probability 
densities. Hypothesis testing and determining confidence sets for PDs 
are discussed in \cite{Chazal2014, Blumberg2014, Chazal2014a, Robinson2017, 
Fasy2014}. One of the main motivations to establish statistical methods for 
hypothesis testing and estimating confidence sets for PH is 
to  distinguish topologically important features from noise. 
% 	 In the context of persistent homology, the \emph{hypothesis test} 
%indicates the quantitative assessment of the computed persistent homology of a 
%dataset sampled from a measure metric space with respect to the true 
%persistent 
%homology. The  estimation of  \emph{confidence sets} of persistence diagrams 
%mainly focuses on differentiating topological noise from topological signals. 
% \cite{Blumberg2014} develops a probability distribution of persistent 
% homological invariant constructed from  subsamples of a fixed size addresses the problem of statistical inference through a variety of 
% test statistics obtained from a pertinent empirical distribution.  
The authors in \cite{Chazal2014} analyze a statistical model for PDs 
obtained from the level set filtration  of a density estimator 
by making use of the bottleneck stability theorem.  
% All of these approaches either 
% rely on a nonparametric permutation test or  reduce 
% persistence diagrams to a summary statistic and  make use of bootstrapping.  
Subsampling either a dataset or its PD to compute statistics of the 
subsamples and to estimate confidence sets of PDs is proposed 
in \cite{Fasy2014}.  
% The authors of \cite{Fasy2014} provide a bound on the 
% bottleneck distance by means of Hausdorff distance and in turn derive the 
% confidence set for PDs. 
% Another article that thoroughly explores the limiting theorems and confidence 
% sets  for PD is \cite{Chazal2018}. 
% The authors argue that in the presence of noise and outliers the   empirical 
%distance function used in several works for confidence sets estimations fails. 
Distance functions based on distance-to-measure and kernel density 
estimation are considered, and the limiting theorem of the empirical distance-to-measure 
depending on the quantile function of the push forward probability is derived in \cite{Chazal2018}. 
%     Hence they consider 
%  distance function based on distance-to-measure (DTM) along with kernel 
%density estimator. The major contributions of this work include estimation of 
%asymptotically valid confidence bands for  DTM, and limiting behavior of 
%kernel 
%distance, and estimation of confidence bands.  The limiting theorem of the 
%empirical DTM depends on the quantile function of the push forward probability 
%defined in \cite{chazal2015a} and its differentiability and uniform modulus of 
%continuity property.  
% As opposed to the work of \cite{Fasy2014}, the algorithm proposed by 
% \cite{Chazal2018} finds different confidence bands for different  homological 
% features. 
The work in \cite{Adler2017, Adler2019} develops a parametric approach based on a Gibbs 
measure that takes the interaction between points in a PD into 
consideration to simulate PDs through Markov chain Monte Carlo (MCMC) 
sampling; the MCMC sampling method therein
assumes a fixed number of points per PD. 

We develop a model that defines PDs as spatially inhomogeneous pairwise interacting 
point processes (PIPPs). 
Typically, the majority of the points in a PD 
are located near the birth axis;
moreover, the topologically significant 
points are fewer in number, lie in the upper portion of the {diagram},  and may be separated from each other. To this end, we 
consider a spatially inhomogeneous model to stochastically treat  the location of 
points in PDs. In particular, we use a Voronoi partition model
to define the spatial density of points in a PD,
% \textcolor{blue}{together with dummy points}
assigning 
higher weights to topologically prominent points in the PD.

Our work proposes a method based on pseudo-likelihood maximization
for estimating the PIPP-based model parameters, and 
%
% \textcolor{blue}{The purpose of the dummy points is to facilitate the assignments of higher weights to the significant points. This is achieved by distributing the dummy points throughout the window so that they are sufficiently dense where  significant points are not located.}
%
% The persistence coordinates of points in a PD indicate the 
% prominence level of corresponding topological features. For example, 
% points with brief persistence may result from noise, whereas 
% points with longer persistence may represent important topological features. The pairwise 
% distance between points in a PD contains information about the underlying dataset.
% A step interaction 
% function with parameters depending on the pairwise distance between points in a %
% PD  is capable of extracting the level of similarity in the corresponding 
% topological features and consequently  becomes an appropriate choice of 
% interaction function.
% Our work proposes a method based on pseudo-likelihood maximization
% for estimating the PIPP-based model parameters. %%
%
develops a
reversible jump MCMC
(RJ-MCMC) 
sampling method to generate random PDs. This method allows 
addition, removal, and relocation of points. Due 
to allowing addition and removal of points, the sampling process is 
trans-dimensional. Our RJ-MCMC sampler traverses the state space of PDs more effectively than existing sampling schemes with regards to capturing topological features (see Section \ref{sec:motivating example}).

RJ-MCMC provides a setting for allowing statistical 
inference related to hypothesis testing and sensitivity analysis.  We provide two examples,
one based on a synthetic dataset as a proof-of-concept, and
one based on a real dataset from materials science
to study the processing-structure-property relationship of AuSS
via hypothesis testing. 

 To summarize, the PIPP model and the RJ-MCMC algorithm make up the RPDG framework, whose main contributions  are the following:
\begin{enumerate}
	\item \textit{A novel PIPP model based on pairwise interactions of PD points},
	which captures the spatial structure of PDs.
	\item  \textit{A novel RJ-MCMC algorithm for sampling PDs based on their PIPP representation}.
	The RJ-MCMC algorithm is flexible enough to accommodate the randomness in 
	the location of points and in the number of points.
	\item 
	\textit{An application of the RPDG in a setting with limited amount of data} to explore processing, microstructure, and property relationships of nano-grained materials.
\end{enumerate}

This paper is organized as follows. 
Section \ref{sec:prelm} provides a brief overview of PDs and 
PIPPs. 
In Section
\ref{sec:main}, we introduce RPDG; in 
Section \ref{sec:model}, we establish the PIPP model for PDs;
in Section \ref{sec:parameter},
we outline parameter estimation for this model; in 
Section \ref{sec:sampling},
we construct the RJ-MCMC algorithm for PD sampling. RPDG is demonstrated and compared to an alternative 
method  in Section
\ref{sec:motivating example}. The performance of our proposed algorithm on AuSS structured materials data is evaluated in Section \ref{sec:application}.
%  To assess the capability of our Bayesian method, we investigate a materials 
%science problem  in Section \ref{sec:sampling example_3}. 
Conclusions are stated in Section \ref{sec:conclusion}.
A proof of Proposition \ref{prop_acc} and details about the design of our RJ-MCMC algorithm
are available in the appendix.

\section{Background}
\label{sec:prelm}

This section outlines
the background required to establish our model of PDs and how to sample PDs based on it. 
Section \ref{subsec:persistence diagram} briefly reviews the  
construction of PDs,
Section \ref{subsec:pipp} provides the basics of PIPPs,
and Section \ref{subsec:rjmcmc_motivation}
motivates the construction of the proposed RJ-MCMC algorithm
for sampling PDs.

\subsection{PDs}
\label{subsec:persistence diagram}

% In this section
We briefly review two frequently used filtration techniques to generate PDs,
{namely filtrations} from point clouds or from functions.
Although we focus on these two types of filtration, RPDG could be generalized
to other filtration techniques for PD generation.

% 	\begin{definition}
% The convex hull of a finite set of points $\{x_i\}_{i=1}^n$ is given by 
%$\sum_{i=1}^n \alpha_i x_i$, where $\alpha_i \geq 0$ for all $i$ and 
%$\sum_{i=1}^n \alpha_i =1$.
% \end{definition}

% \begin{definition}
% The set of points $\{x_i\}_{i=1}^n$ is affinely independent if whenever $ 
%\sum_{i=1}^n \alpha_i x_i = 0$ and $\sum_{i=1}^n \alpha_i = 0$, then $\alpha_i 
%= 0$ for all $i$.
% \end{definition}

% \begin{definition}
%     A $k$-simplex is the convex hull of an affinely independent point set of 
%cardinality $k+1$. The convex hull of a nonempty subset of the $k$ points in a 
%$k+1$ simplex is called a face of a simplex.
%     \end{definition}

%     \begin{definition}
%       A simplicial complex $\sigma$ is a collection of simplices such that for
%         every set $A$ in $\sigma$ and every nonempty set $B \subset A$, we 
%have that $B$ is in $\sigma$.
%         \end{definition}

% To approximate the shape of the underlying point cloud $X$, we use a sequence 
%of  Vietoris-Rips complexes.

\subsubsection{Filtration from point clouds} \label{subsec:vr}

\begin{figure}[!ht]
	\centering
	\subfloat[]{\includegraphics[width = 0.2\textwidth]{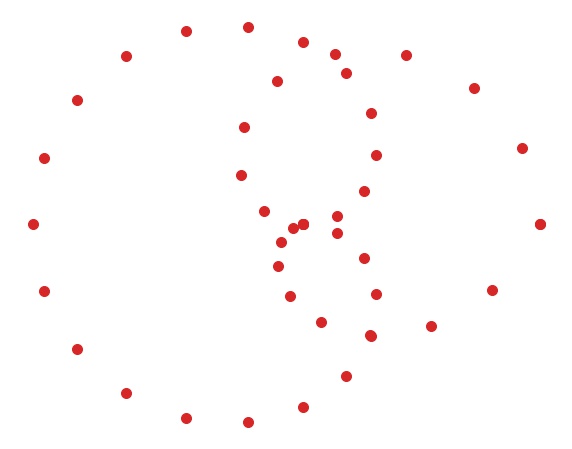}\label{fig:VR_a}}\hspace{0.1in}
	\subfloat[]{\includegraphics[width = 0.2\textwidth]{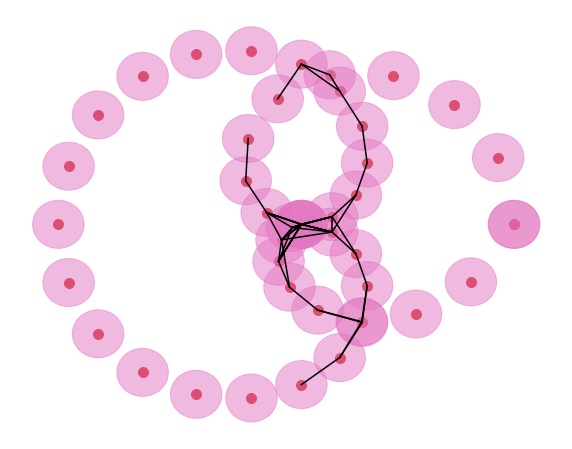}\label{fig:VR_b}}\hspace{0.1in}
	\subfloat[]{\includegraphics[width = 0.2\textwidth]{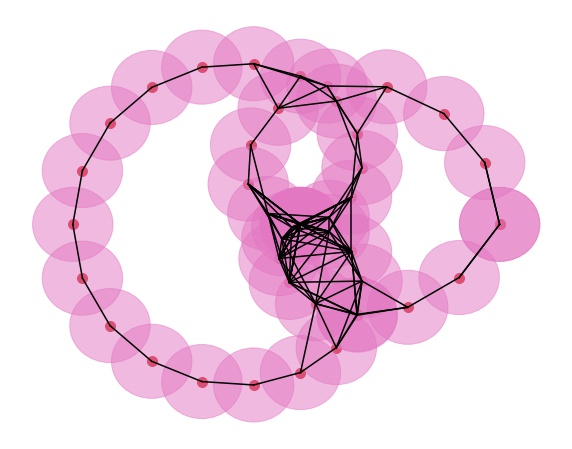}\label{fig:VR_c}}\hspace{0.1in}
	\subfloat[]{\includegraphics[width = 0.21\textwidth]{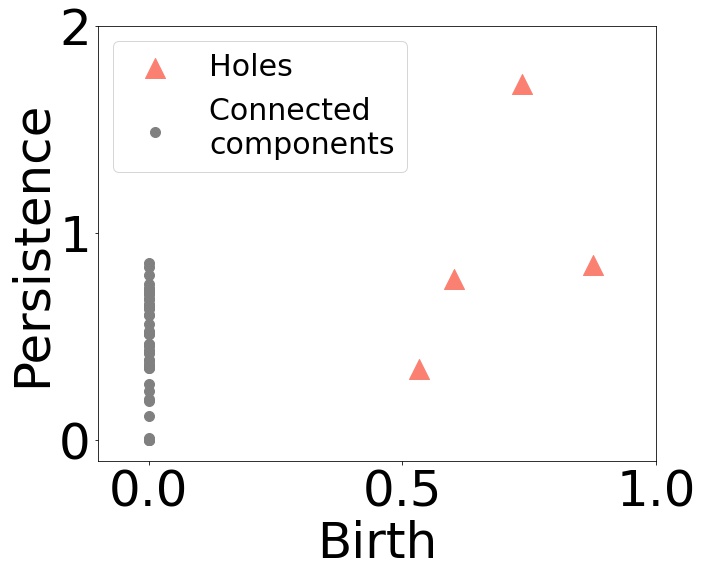}\label{fig:VR_d}}
	\caption{{(a) A point cloud with 45 (red) points.
	(b) A Vietoris-Rips complex of the point cloud in (a) for radius $\zeta_i$.
	(c) Another Vietoris-Rips complex of the point cloud in (a), with radius $\zeta_j > \zeta_i$.
	(d) A tilted PD for connected components and holes
	associated with a sequence of Vietoris-Rips complexes.
	}}
	\label{fig:VR}
%	\vspace{-0.3in}
\end{figure}

{
The Vietoris-Rips filtration is introduced below, including its building blocks.
An illustration of Vietoris-Rips filtration is displayed in Figure~\ref{fig:VR}.}

% \begin{definition}
% 	Let $X=\{ x_1, \dots, x_n \}$ be a point cloud in $\R^{\mathbbm{d}}$. The 
% 	Vietoris-Rips complex VR$(\epsilon)$ of $X$ is defined to be the simplicial 
% 	complex $V_{\epsilon}(X)$ satisfying
% 	$[x_{i_{1}},\dots,x_{i_{l}}] \in V_{\epsilon}(X)$ if and only if 
% 	$\diam(x_{i_{1}},\dots,x_{i_{l}}) < \epsilon$.
% 	Given a nondecreasing sequence $\{\epsilon_{n}\} \in \R^{+} \cup \{0\}$  with 
% 	$\epsilon_{0} = 0$, we define its Vietoris-Rips filtration as
% 	$\{V_{\epsilon_{n}}(X) \}_{n \in \N}$.
% \end{definition}
% 
% 
%

% An illustration of Vietoris-Rips complexes is shown in Fig. \ref{fig:VR}. 

\begin{definition}{
	A $\psi$-dimensional collection of data $\{v_{0},\dots,v_{
	\tau}\} \subset \R^{\psi} \setminus \{0\} $ is said to be geometrically independent if for any set $t_{i} \in \R$ with $\sum_{i = 0}^{
	\tau} t_{i} = 0$, the equation
	$\sum_{i=0}^{\tau}t_{i}v_{i} = 0$
	implies that
	$
	t_{i} = 0$  for all $i\in \{ 0,\dots,\tau \}.
	$}
\end{definition}
%-----------------

% It can be verified that $v_{0},\dots,v_{n}$ are geometrically independent if and only if the vectors $v_{1} - v_{0}, \dots, v_{n} - v_{0}$ are linearly independent \cite{Naber1980}. Requiring all of the $v_{i}$'s to be nonzero ensures that one element sets are always geometrically independent \cite{munkres_elements_of_alg_top}.
%--------------
\begin{definition}
\label{def:simplex}
{
A $\kappa-$simplex, is a collection of $\kappa+1$ geometrically independent elements  with their convex hull
$$[v_{0},\dots,v_{\kappa}] = \Big\{\sum_{i = 0}^{\kappa}\omega_{i}v_{i} : \sum_{i = 0}^{\kappa}\omega_{i} = 1\Big\}.$$
We say that the vertices $v_{0},\dots,v_{\tau}$ span the $\kappa-$dimensional simplex, $[v_{0},\dots,v_{\kappa}]$.
The faces of a $\kappa-$simplex $[v_{0}, \dots, v_{\kappa}]$, are the $(\kappa-1)-$simplices spanned by subsets of $\{v_{0},\dots, v_{\kappa}\}$. }
%We denote the collection of faces by $\mathcal{F}_{[v_{0},\dots,v_{k}]}$. 
\end{definition}
%-----------------

%\begin{remark}
%	One can easily deduce from Definition \ref{def:simplex} that simplices are invariant under permutation of vertices. Hence the set $\mathcal{F}_{[v_{0},\dots,v_{k}]}$ contains $\binom{k+1}{k} = k+1$ unique elements.
%\end{remark}

%---------------
\begin{definition}
	\label{simpcomp}{
	A simplicial complex $S_c$ is a collection of simplices satisfying two conditions: (i) if $\xi \in S_c$, then all faces of $\xi$ are also in $S_c$, and (ii) the intersection of two simplices in $S_c$ is either empty or contained in $S_c$. }
	%We denote the the collection of $k$-simplices contained in $\K$ by $\K^{[k]}$
\end{definition}

{Given a point cloud, $V$, our goal is to construct a sequence of simplicial complexes that reasonably approximates the underlying shape of the data.  We accomplish this by using the Vietoris-Rips filtration.}
  %as the geometric complexes are typically used for applications of persistent homology in data analysis \cite{1307.6188, edelsbrunner2010computational, Marchese2018, Venkataraman2016}.

\begin{definition}
	\label{ripsfilt}
{	Let $V=\{ v_0,\ldots,v_{\tau} \}$ be a point cloud in $\R^{\psi}$ and $\zeta>0$. The Vietoris-Rips complex of $V$ is defined to be the simplicial complex $\mathcal{V}_{\zeta}(V)$ satisfying
	$[v_{i_{1}},\dots,v_{i_{l}}] \in \mathcal{V}_{\zeta}(V)$ if and only if $\diam(v_{i_{1}},\dots,v_{i_{l}}) < \zeta$.
	Given a nondecreasing sequence $\{\zeta_{\tau}\} \in \R^{+} \cup \{0\}$  with $\zeta_{0} = 0$, we denote its Vietoris-Rips filtration by
	$\{\mathcal{V}_{\zeta_{\tau}}(V) \}_{\tau \in \N}$.}
\end{definition}

{A PD $\D$  is a multi--set of points in $ \W \times \{0,1, \dots, \psi-1 \}$, where 
\begin{equation}
\label{eq:wedge}
\mathbb{W} = \{d = (\beta,\delta - \beta)  \in \R^{2} \mid \beta, \delta -\beta \geq 0 \}.
\end{equation}
For a fixed dimension $\kappa  = 0 , \dots, \psi-1$, each element $(\beta, \delta-\beta)$  represents a homological feature of dimension $\kappa$ that appears at scale $\beta$ during a Vietoris-Rips filtration, and disappears at scale $\delta$. In other words, the homological feauture $(\beta, \delta-\beta)$ is a $\kappa-$dimensional hole that persists $\delta - \beta$.  Features with $\kappa=0$ correspond to connected components, $\kappa=1$ to loops, and $\kappa=2$ to voids. An illustration of Vietoris-Rips filtration and an example of a PD is given in Figure \ref{fig:VR}.}

\subsubsection{Filtration from functions}  \label{subsec:sl}

For a real number $\epsilon$, the sublevel set $S_{\epsilon}$ of a function,
{$f:\mathbb{R}\rightarrow\mathbb{R}$},
is defined as
$S_{\epsilon} = f^{-1}((-\infty,\epsilon])$.
A collection $\{S_{\epsilon}:\epsilon\in\mathbb{R}\}$ of sublevel sets of $f$
is called a sublevel set filtration of $f$.
A sublevel set filtration tracks the evolution of connected components,
that is of zero-dimensional homological features, as ${\epsilon}$ increases. As all of the sublevel sets $S_{\epsilon}$ are either empty or a union of intervals, we can extract information about the connectivity of the sets $S_{\epsilon}$, which in turn provides the number of connected components.
% As $\epsilon$ increases, the number of connected components of $S_{\epsilon}$ remains unchanged except 
% when it passes through a critical point.
We record the value of $\epsilon$ (local minimum of $f$) at which a given connected component is born,
and the value of $\epsilon$ (local maximum of $f$) at which
the connected component
disappears by merging with a pre-existing connected component. According to the elder rule \cite{Edelsbrunner2010}, whenever two connected components merge, the one born later disappears while the one born earlier persists. Once $\epsilon$ takes the maximum value $\max f(t)$, all the sublevel sets merge into a single connected component.
% , and we terminate the procedure.

\begin{figure}[!ht]
	\centering
    \subfloat[]{\includegraphics[width=0.307\textwidth]{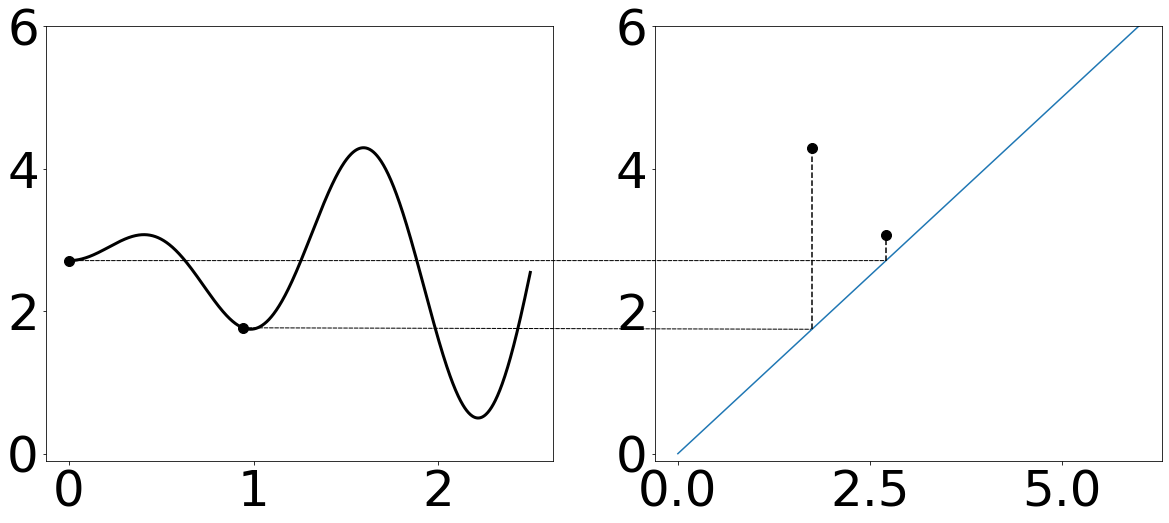}}
    \subfloat[]{\includegraphics[width=0.165\textwidth]{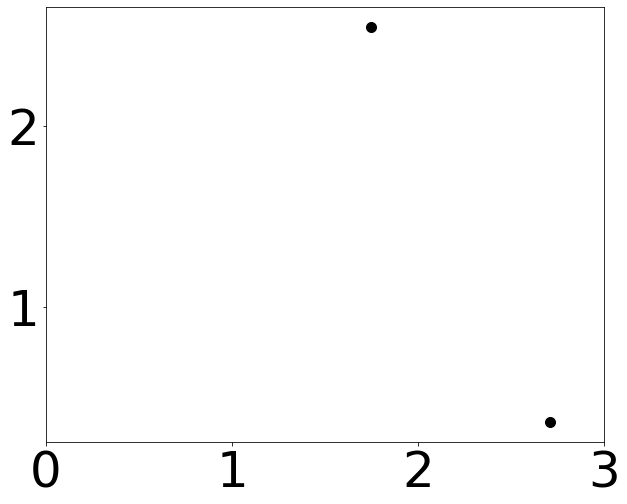}}
	\caption{(a) A continuous function and the PD of its sublevel set filtration.
	% the sublevel set filtration of the function. %the red dot represents the infinite generator, 
	(b) The tilted PD obtained from the sublevel set filtration.}
	\label{fig:sl}
%	\vspace{-0.3in}
\end{figure}

For every connected component that arises in the filtration, we track the points $({\beta},{\delta}) \in \mathbb{R}^{2}$, where $\beta$ is the value of $\epsilon$ at which the connected component is born and $\delta$ is the value of $\epsilon$ at which it disappears,  and call the resulting collection a PD. Similarly to Section \ref{subsec:vr}, one may apply the linear transformation $d = (\beta,\delta-\beta)$  and consider the associated wedge $\mathbb{W}$. 
%Hence the persistence diagram (PD) is a multiset of points $D=\{\zeta\}=\{(b,d)\}$  in  $\widetilde{\W} := \{(b,d) \in \R^{2} | \,\, d \geq b \geq 0\}$. 
% We then apply the linear transformation $d = (\beta,p)=T(\beta,\delta) = (\beta-\min(\beta),\delta-\beta)$ to each point in our PDs. We refer to the resulting coordinates as birth ($\beta$) and persistence ($p$), respectively, in  $\W := \{d =(\beta,p) \in \R^{2} | \,\, \beta,p \geq 0\}$ and call this transformed PD a tilted representation. 
An illustration of a sublevel set filtration of a function
and of the tilted PD based on the filtration are shown in Figure \ref{fig:sl}.

%\vspace{-0.1in}

\subsection{PIPPs}
\label{subsec:pipp}

Here, we present the components that we later employ in Section~\ref{sec:main} to
construct our RPDG framework.
% In particular,
Section~\ref{sec:pipp_density} states
the definition of a  pairwise interacting point
process (PIPP), including
the probability density function (pdf) of a set of points in the PIPP.
Sections~\ref{sec:spatial_density} and~\ref{sec:interaction} provide
a spatial {pattern} and a pairwise interaction function, respectively, that
can be used to fully specify the pdf of a set of ponts in a PIPP.
Our PIPP model for PDs (Section~\ref{sec:model}) and
our RJ-MCMC algorithm that samples PDs using our model (Section~\ref{sec:sampling})
are built upon the PIPP density specified across
Sections~\ref{sec:pipp_density},~\ref{sec:spatial_density} and~\ref{sec:interaction}.
The PIPP pseudolikelihood of Section~\ref{sec:pipp_pseudolik}
is used for inferring the parameters of our PIPP model for PDs,
as elaborated in Section~\ref{sec:parameter}.
% The PIPP model for PDs of Section X is built upon the density of Section X and X,
% while the RJ-MCMC sampling algorithm of Section X samples for that

\subsubsection{PIPP density}
\label{sec:pipp_density}

{One of our central motivations
is to capture the local and global spatial features of the distribution of points in a PD.
A PIPP is a Gibbs point process with density function determined by a first and second order potential function \cite{Baddeley2000}.
In the context of a PD,
the first order potential function captures
the spatial density of points in the PD
and the second order potential function determines
interactions between all possible pairs of points.}

%Let $(\mathbb{W}, \mathcal{W}, \lambda)$ be a measure space, where $\mathbb{W}$ 
%is a bounded region in $\mathbb{R}^2$,
%% where a point process is observed, 
%$\mathcal{W}$ is the Borel $\sigma$-algebra on $\mathbb{W}$ and $\lambda$ is 
%the Lebesgue  measure.
%In Definition~\ref{def:pipp},
%we consider a spatial point process consisting of a set of points
%$\mathbf{x} = \{x_i \in \mathbb{W}:i=1,\ldots,n\}$
%such that the number of points in any region $R \subseteq \mathcal{W}$
%has a Poisson distribution with mean $\lambda(R)$.

% Denote $\mathbb{X}_n$ as the set of all possible configuration of  $n$ points 
%with a default setting of $\mathbb{X}_0 = \{\varnothing\}$. A point process on 
%$\mathbb{W}$ is a random variable on the space $(\mathbb{X}, \mathcal{X}, 
%\mu)$ 
%where $\mathcal{X}$ is the Borel $\sigma$ algebra on $\mathbb{X}$ and $\mu$ is 
%the intensity measure of a homogeneous point process that admits the intensity 
%density $\lambda$ \cite{Daley1988}. 

% We begin by discussing the necessary background for generating pairwise 
%interacting point process models for PDs. 
\begin{definition}[PIPP]
\label{def:pipp}
Let $(\mathbb{W}, \mathcal{W}, \lambda)$ be a measure space, where {$\mathbb{W}$ is the set defined in Equation \eqref{eq:wedge}, and in addition, 
a bounded region of $\mathbb{R}^2$},
% where a point process is observed, 
$\mathcal{W}$ is the Borel $\sigma$-algebra on $\mathbb{W}$, and $\lambda$ is 
the Lebesgue  measure.
A pairwise interacting point process $X$ is a spatial point process on $(\mathbb{W}, \mathcal{W}, \lambda)$ 
with spatial pattern function $s:\mathbb{W} \rightarrow \mathbb{R}^{+} \cup 
\{0\}$ and interaction function $h_{\theta}:\mathbb{W} \times \mathbb{W} 
\rightarrow \mathbb{R}^{+} \cup \{0\}$.
For a set of points $\mathbf{x} = \{x_1, \dots, x_n\}\subseteq\mathbb{W}$ of $X$,
the pdf $f(\mathbf{x}\mid\theta)$ of $\mathbf{x}$ has the form
%\begin{equation}
\begin{align}
\label{eqn:pairwise_int_density}
f(\mathbf{x}\mid\theta)
% =\frac{1}{Z(\theta)}\prod_{i=1}^ns(x_i)\prod_{i<j}h_{\theta}(x_i,x_j)\\
& =\frac{1}{Z(\theta)}\prod_{i=1}^ns (x_i)g(\mathbf{x}\mid\theta),\\ \label{eq:g}
%\end{equation} 
%where 
%\begin{align}
g(\mathbf{x} \mid \theta) &= \prod_{i<j} h_{\theta} (x_i,x_j), 
%\label{eq:Z}
% Z(\theta) &=  \int _{\mathbb{W}}\prod_{i=1}^n s (x_i)
% % \prod_{i<j} h_{\theta} (x_i,x_j)
% g(\mathbf{x} \mid \theta)
% \,d\lambda(\mathbf{x}),
\end{align} 
where $\theta  = (\theta_1, \dots, \theta_k)\in\mathbb{R}^k$ is a vector of parameters, and $Z(\theta) =  \int _{\mathbb{W}}\prod_{i=1}^n s (x_i)
% \prod_{i<j} h_{\theta} (x_i,x_j)
g(\mathbf{x} \mid \theta)
\,d\lambda(\mathbf{x}),$ is the normalizing constant.
\end{definition}

According to Definition~\ref{def:pipp},
a PIPP is a spatial point process.
Thus, the number of points of a PIPP
in any region $R \subseteq \mathbb{W}$
follows a Poisson distribution with mean $\lambda(R)$.
%
% The interaction term $h_{\theta}(x_i,x_j)$ in Equation~\eqref{eqn:pairwise_int_density}
% depends on the Euclidean distance $\|x_i-x_j\|$ between $x_i$ and $x_j$,
% and on parameter $\theta$.
%
The normalizing constant  $Z(\theta)$ is typically intractable,
{i.e. it is not available in closed form or
it is computationally expensive. An example of interaction function $h_\theta$ of Equation \eqref{eq:g} and the associated parameter vector $\theta$ are given in Section~\ref{sec:interaction}. More specifically, see Equation \eqref{pcpi_function}.}

% Equation~\eqref{eqn:pairwise_int_density} can be written as 
% \begin{equation}
% f(\mathbf{x} \mid \theta) = \frac{1}{Z(\theta)} 
% \exp(U_{\theta}(\mathbf{x})),\label{eqn:Gibbs_density}
% \end{equation} 
% where $f(\mathbf{x} \mid \theta)$ takes the form of the pdf of 
% a finite Gibbs point process with  potential function 
% % \vspace{-0.3in}
% \begin{equation} \label{eqn:potential_function}
% U_{\theta}(\mathbf{x})  = \sum_{i=1}^n \ln(s (x_i)) + 
% \sum_{i < j}\ln
% (h_{\theta} (x_i,x_j))
% \end{equation}

\subsubsection{{A spatial pattern function}}
\label{sec:spatial_density}

One way of specifying the spatial pattern function $s$ in
Equation~\eqref{eqn:pairwise_int_density}
is based on the {notion of Voronoi diagrams}.
Along these lines,
we recall what is a Voronoi cell (Definition~\ref{def:dirichlet_tile}),
which constitutes a building block for a
Voronoi diagram (Definition~\ref{def:dirichlet_tessellation}).
Subsequently, we state the {{spatial pattern}}
induced by a Voronoi diagram
(Definition~\ref{def:spatial_density}).

% We next define the notion of Dirichlet tile.
% to describe the spatial density term $s$ of the pdf 
% $f(\mathbf{x}\mid\theta)$ in  Equation \eqref{eqn:pairwise_int_density}. 

\begin{definition}[Voronoi cell]
\label{def:dirichlet_tile}
Let $\{x_1, \dots, x_n\}$ be a set of distinct points in a bounded region
$\mathbb{W}$ of $\mathbb{R}^2$.
The Voronoi cell $T_i,~i=1,\ldots,n$,
associated with $x_i$
% associated with $u_i$
is defined as
\begin{equation*}
T_i =
\{x\in\mathbb{W}:
\|x-x_i\| \le \|x-x_j\|\,\, \forall \,\, j~\mbox{with}~j\neq i\},
\end{equation*}
where $\|\cdot\|$ denotes the Euclidean norm.
\end{definition}

\begin{definition}[Voronoi diagram]
\label{def:dirichlet_tessellation}
Let $\{x_1, \dots, x_n\}$ be a set of distinct points in a bounded region
$\mathbb{W}$ of $\mathbb{R}^2$.
Moreover, let $T_i,~i=1,\ldots,n$,
be the Voronoi cell associated with $x_i$.
The Voronoi diagram
associated with $\{x_1,\ldots,x_n\}$
is defined as the collection 
$\{T_1,\ldots,T_n\}$
of Voronoi cells.
\end{definition}

A Voronoi cell $T_i$ has the property that any point in the interior of 
$T_i$ is closer to point $x_i$ than to any other point $x_j,~j\neq i$
\cite{Okabe2000}.
{
RJ-MCMC for PDs, as discussed in Section~\ref{sec:sampling},
samples almost surely from the Voronoi cell interiors.}
% Without loss of generality,
% we assume that the cells in a Voronoi diagram are pairwise disjoint.
% we assume that the cell boundaries in a Voronoi diagram
% are allocated among the cells
% so that the cells are pairwise disjoint.

\begin{definition}[Spatial pattern induced by a Voronoi diagram]
\label{def:spatial_density}
Let $\{T_1,\ldots,T_n\}$
be the Voronoi diagram associated with a set of points $\{x_1, \dots, x_n\}$
in a bounded region
$\mathbb{W}$ of $\mathbb{R}^2$.
Let $A_i,~i=1,\ldots,n$,
be the area of Voronoi cell $T_i$.
The {{spatial pattern}} function
$s:\mathbb{W} \rightarrow \mathbb{R}^{+} \cup \{0\}$
induced by $\{T_1,\ldots,T_n\}$
is defined as
\begin{equation}
\label{eq:s_v}
s(x) = \sum_{l=1}^{n} A_l\mathbbm{1}_{\{x\in T_l\}},
\end{equation}
where $x\in \mathbb{W}$, and
$\mathbbm{1}_{\{\cdot\}}$
denotes the indicator function.
\end{definition}

% As the cells of Voronoi diagram
% $\{T_1,\ldots,T_n\}$
% are pairwise disjoint,
% it follows that any $x\in\mathbb{W}$
% belongs to exactly one cell $T_i$.
% So, for any $x\in\mathbb{W}$,
% the sum in Equation~\eqref{eq:s_v}
% collapses to $s(x)=A_i$,
% where $A_i$ is the area
% of the unique Voronoi cell $T_i$ to which $x$ belongs.

% Consider a set $\{x_1, \dots, x_n\}$ of points of a PIPP.
Consider a PIPP with points $\{x_1, \dots, x_n\}$. 
The PIPP {{pattern function}} $s$ in Equation~\eqref{eqn:pairwise_int_density}
can be set via
the Voronoi diagram associated with points $\{x_1, \dots, x_n\}$.
For a PIPP point $x_i\in T_i,~i=1,\ldots,n$,
it follows from Equation~\eqref{eq:s_v}
that $s(x_i)=A_i$, where $A_i$ is the area of cell $T_i$.

% For a set $\{x_1, \dots, x_n\}$ of points in a PIPP,
% the spatial density $s$

% Consider a set $\{x_1, \dots, x_n\}$ of points in a PIPP.
% The PIPP spatial density 

% Subsequently, we state the spatial density
% for a set of points in a PIPP induced by a Voronoi diagram
% (Definition~\ref{def:spatial_density}).

\subsubsection{A pairwise interaction function}
\label{sec:interaction}

The interaction term $h_{\theta}(x_i,x_j)$ in Equation~\eqref{eq:g}
is chosen typically so that it depends on the Euclidean distance $\|x_i-x_j\|$
% between $x_i$ and $x_j$,
and on parameter $\theta$.
The piece-wise constant pairwise interaction function (Definition~\ref{def:step_int})
can be used 
in Equation~\eqref{eq:g}
as the interaction function $h_{\theta}$
for pairs of points in a PIPP.
This is also known as the multi-scale generalization of the Strauss interaction \cite{strauss1975}.
%which is a frequently used interaction function in the literature.
% The Strauss interaction function has a constant value for pairwise distances less than a threshold, and 1 otherwise.
% The PCPI function relies on distance cut-offs to model the interaction between points in a point process.
%PCPI is capable of 
%extracting the level of similarity in the corresponding topological features. 
%We present an example to exhibit this phenomenon in Section 
%\ref{sec:motivating example}.         

\begin{definition}[Piece-wise constant pairwise interaction function]
\label{def:step_int}
Let $\mathbb{W}$ be a bounded region in $\mathbb{R}^2$.
The piece-wise constant pairwise interaction function
$h_{\theta}:\mathbb{W} \times \mathbb{W} 
\rightarrow \mathbb{R}^{+} \cup \{0\}$
% $h_{\theta}(x_i,x_j)$ for any pair of points $(x_i,x_j) \in X$
is defined as
% by the indicator function  $\mathbbm{1}_{\{r_{l-1} < \|x_i-x_j\| <r_l\}},~l=1, \dots, k$, and the parameters $\theta  = \{\theta_l \in \mathbb{R} \}_{l=1}^k$ as
%	\vspace{-0.4in}
{{
\begin{equation}
\label{pcpi_function}
h_{\theta}(x,z) = \exp\left(
\sum_{l=1}^k \theta_l
% I_l(\|x_i - x_j\|) \right).
{\mathbbm{1}_{\{r_{l-1} < \|x-z\| \le r_l\}}} \right),
\end{equation}
}}
where
$(x,z) \in \mathbb{W} \times \mathbb{W}$,
% $I_l(\|x_i-x_j\|) = \mathbbm{1}_{\{r_{l-1} < \|x_i-x_j\| <r_l\}}$,
$r_l\in\mathbb{R}$ for $l=0,1,\ldots,k$, satisfying
$0 =r_0<r_1< \ldots < r_k$,
and $\mathbbm{1}_{\{\cdot\}}$ denotes the indicator function.
The vector $\mathbf{r} = (r_1, \dots ,r_k)$ is called the vector of jump points.
\end{definition}

% \begin{definition}[Piece-wise constant pairwise interaction function]
% \label{def:step_inta}
% Let $\mathbf{x} = \{x_1, \dots, x_n\}$
% be a set of points of a PIPP $X$ with density
% $f(\mathbf{x} \mid \theta)$,
% where $\theta  = (\theta_1, \dots, \theta_k)\in\mathbb{R}^k$.
% The piece-wise constant pairwise interaction function
% % $h_{\theta}(x_i,x_j)$ for any pair of points $(x_i,x_j) \in X$
% is defined as
% % by the indicator function  $\mathbbm{1}_{\{r_{l-1} < \|x_i-x_j\| <r_l\}},~l=1, \dots, k$, and the parameters $\theta  = \{\theta_l \in \mathbb{R} \}_{l=1}^k$ as
% %	\vspace{-0.4in}
% \begin{equation}
% \label{pcpi_function}
% h_{\theta}(x_i,x_j) = \exp\left(
% \sum_{l=1}^k \theta_l
% % I_l(\|x_i - x_j\|) \right).
% {\mathbbm{1}_{\{r_{l-1} < \|x_i-x_j\| <r_l\}}} \right),
% \end{equation}
% where
% $x_i,x_j \in X$,
% % $I_l(\|x_i-x_j\|) = \mathbbm{1}_{\{r_{l-1} < \|x_i-x_j\| <r_l\}}$,
% $r_l\in\mathbb{R}$ for $l=0,1,\ldots,k$, satisfying
% $0 =r_0<r_1< \ldots <r_k$,
% and $\mathbbm{1}_{\{\cdot\}}$ denotes the indicator function.
% The vector $r = (r_1, \dots ,r_k)$ is called the vector of jump points.
% \end{definition}

The interaction function of Equation~\eqref{pcpi_function} 
is a piece-wise constant function, whose value
% relies on distance cut-offs $r$
% to model the interaction between points in a point process.
% In Equation~\eqref{pcpi_function},
% the value
$h_{\theta}(x,z)$ depends only on the distance between $x$ and $z$;
if {{$r_{l-1} < \|x-z\| \le r_l$}}, then
%More specifically,
$h(x,z) = \exp( \theta_l)$. 
We thus interpret the jump points $r$
as points of discontinuity of $h_{\theta}$, {{and the parameter vector $\theta$ as a set of weights that determines how important is the interaction among PD points.}}

% In particular, to model the interaction between the pair $(x_i,x_j)$ the PCPI function relies on the number of points $x_i \in \mathbf{x}$ whose distance from $x_j$ for $i \neq j$ is within the interval $(r_{l=1}, r_l)$.   

% In practice, the persistence coordinates of points in a PD indicate the 
%prominence level of corresponding topological features. For example, typically 
%points with brief persistence may result from noise whereas, longer 
%persistence 
%points represent important topological features. Furthermore, the pairwise 
%distance between points in a PD contains information about the underlying data 
%set. Nearby points usually share similar topological features and points that 
%are further away mostly represent contrasting features. To this end, a step 
%function with parameters depending on the pairwise distance between points in 
%a 
%PD is 

\subsubsection{PIPP log-pseudolikelihood}
\label{sec:pipp_pseudolik}

The density
$f(\mathbf{x} \mid \theta)$
given by Equation~\eqref{eqn:pairwise_int_density}
% of a PIPP
% with points $\mathbf{x}$
can be employed as a likelihood function.
% of the PIPP.
A PIPP likelihood function $f(\mathbf{x} \mid \theta)$,
as specified by Equation~\eqref{eqn:pairwise_int_density},
is computationally expensive,
since the normalizing constant $Z(\theta)$ is intractable.
A pseudolikelihood
can be used as a computationally feasible approximation of 
$f(\mathbf{x} \mid \theta)$.
According to~\cite{Baddeley2000},
we state the pseudolikelihood 
of a PIPP (Definition~\ref{def:pseudolikelihood})
based on the conditional intensity of the PIPP (Definition~\ref{def:cond_intensity}).

%  The pseudolikelihood definition 
% is based on the conditional intensity $\mathcal{I}(u,\mathbf{x})$ (see Definition \ref{def:conditional intensity} below), which is the 
% conditional probability that a point process $X$ has a point $u \in \mathbb{W}$  given that $X$  consists of the points $\mathbf{x} = (x_1, \ldots, x_n)$.     

\begin{definition}[PIPP conditional intensity]
\label{def:cond_intensity}
Let $\mathbf{x}=\{x_1,\ldots,x_n\}$
be a set of points of a PIPP $X$
with density
$f(\mathbf{x} \mid \theta)$,
where $\theta  = (\theta_1, \dots, \theta_k)\in\mathbb{R}^k$.
The conditional intensity of $X$ is defined as
\begin{equation}
\mathcal{I}(u,\mathbf{x}) = \left\{
\begin{array}{ll}
\frac{f(\mathbf{x} \cup {u}\mid \theta)}{f(\mathbf{x}\mid \theta)} & u \notin \mathbf{x} \\
\frac{f(\mathbf{x} \mid \theta)}{f(\mathbf{x}\setminus {u}\mid \theta)} & u \in \mathbf{x} .
\end{array} 
\right. 
%\mathcal{I}(u,\mathbf{x}) =
% \frac{f(\mathbf{x} \cup {u})}{f(\mathbf{x})} \,\,\, 
%	\text{if} \,\,\, u \notin \mathbf{x}
% \end{equation}
% 	or \[\mathcal{I}(x_i,\mathbf{x}) = \frac{f(\mathbf{x} )}{f(\mathbf{x} \setminus 
% 	{x_i})} \,\,\, \text{if} \,\,\, x_i \in \mathbf{x}\]
\end{equation}
\end{definition}

% As seen from Definition~\ref{def:cond_intensity},
For a PIPP $X$ on $(\mathbb{W}, \mathcal{W}, \lambda)$,
the conditional intensity $\mathcal{I}(u,\mathbf{x})$
is the conditional probability that $X$
has a point $u$ in $\mathbb{W}$ given 
that $X$ consists of $\mathbf{x}$.
For the PIPP density of Equation \eqref{eqn:pairwise_int_density},
the conditional intensity takes the form
\begin{equation}
\label{eqn:conditional intensity}
\mathcal{I}(u,\mathbf{x}\mid \theta) =
%\exp \left((1,\theta)^T
%(S(u),\sum_{i=1,x_i \neq u}^n H(u,x_i))\right),
% \exp \left(S(u)+\sum_{\mathclap{\substack{i=1\\x_i \neq u}}}^n
% \sum_{l=1}^{k}
% \theta_{l}I_l(\|u-x_i\|) )
%
s(u) \prod_{\mathclap{\substack{i=1\\x_i \neq u}}}^n 
h_{\theta}(u, x_i),
\end{equation}
%where $\theta = (\theta_1, \dots, \theta_k)$ is the parameter vector, and
%$h_{\theta}(u, x_i)$ is given by Eq (6)...
%, and
%\begin{align*}
%H(u,x_i) &= (I_1(\|u-x_i\|), \dots, I_k(\|u-x_i\|) ),\\
% $I_l(\|u-x_i\|) = \mathbbm{1}_{\{r_{l-1} < \|u-x_i\| <r_l\}}$.
%~l=1, \dots, k.
%\end{align*}
 %Hereafter we write  $\theta = ( \theta_1, \dots, \theta_k)$ interchangeably by abusing notation. 
\begin{definition}[PIPP log-pseudolikelihood]
\label{def:pseudolikelihood}
The log-pseudolikelihood of a PIPP $X$ with conditional intensity 
$\mathcal{I}(u,\mathbf{x}\mid\theta)$ is defined as
\begin{align} 
\nonumber&\log \tilde{L}(\theta\mid \mathbf{x}) =\\&
\sum_{i=1}^n \log \mathcal{I}(x_i, \mathbf{x}\mid\theta) 
-\int_{\mathbb{W}}
\mathcal{I}(u,  \mathbf{x}\mid\theta)du.\label{eqn:logpseudolikelihood}
\end{align}
% \begin{align} 
% \nonumber &\tilde{L}(\theta\mid \mathbf{x}, R) =\\&
% \prod_{x_i \in R} \mathcal{I}(x_i, \mathbf{x}\mid\theta) 
% \exp\left(-\int_{R}\mathcal{I}(u,  \mathbf{x}\mid\theta)du\right), \label{eqn:pseudolikelihood}
% \end{align}
% for any region $R \subseteq \mathbb{W}$.
\end{definition}

% Hence, the log-pseudolikelihood takes the form
% \begin{align} 
% \nonumber&\log \tilde{L}(\theta\mid \mathbf{x}, R) =\\&
% \sum_{i=1}^n \log \mathcal{I}(x_i, \mathbf{x}\mid\theta) 
% -\int_{R}\mathcal{I}(u,  \mathbf{x}\mid\theta)du.\label{eqn:logpseudolikelihood}
% \end{align}

If the conditional intensity $\mathcal{I}$
in Equation~\eqref{eqn:logpseudolikelihood}
employs the piece-wise constant pairwise interaction function,
then the PIPP log-pseudolikelihood can be approximated
by the Berman-Turner device~\cite{Berman1992,Baddeley2000}.
The PIPP log-pseudolikelihood approximation based on the
Berman-Turner device is given by
\begin{align} 
\nonumber&\log \tilde{L}(\theta\mid \mathbf{x}) \approx\\&
\sum_{j=1}^m
% \left(
\mathbbm{1}_{\{u_j\in\mathbf{x}\}}
\log\mathcal{I}(u_j,\mathbf{x}\mid \theta)-
w_j\mathcal{I}(u_j,\mathbf{x}\mid \theta),
% \right),
\label{eqn:bt_logpseudolik}
\end{align}
where
$\{u_1,\ldots,u_m\}$ are points in $\mathbb{W}$
such that
$\{x_1,\ldots,x_n\}\subseteq\{u_1,\ldots,u_m\}$,
and $\{w_1,\ldots,w_m\}$ are positive weights
summing to the area of $\mathbb{W}$.

% The pseudolikelihood of a pairwise interacting point process representation 
%of PDs entails the estimation of unknown parameters. 

\subsection{Motivation for RJ-MCMC}
\label{subsec:rjmcmc_motivation}

The idea of sampling PDs to perform statistical inference
was introduced by \cite{Adler2017} and was improved in \cite{Adler2019}.
In \cite{Adler2017, Adler2019},
a Metropolis-within-Gibbs (MWG) algorithm was constructed to sample PDs
by randomly relocating PD points.
However, relocating points of a PD is not the only move one may consider.
For example, one may observe more or fewer number of points in a PD
depending on the noise level in the data.

In this paper, we construct an RJ-MCMC algorithm to sample PDs.
RJ-MCMC is an MCMC algorithm developed by \cite{Green1995}
to enable simulation from a distribution on spaces of varying dimensions.
In the context of PDs, RJ-MCMC enables additional types of stochastic moves
in comparison to the MWG approach of \cite{Adler2017, Adler2019}.
More specifically, we develop an RJ-MCMC algorithm
which generates new PDs not only by relocating points,
but also by adding or removing points.
The addition or removal of points generates PDs with different number of points.
Our RJ-MCMC scheme solves the problem of sampling
from a distribution of PDs with varying number of points,
in contrast to the MWG scheme of \cite{Adler2017, Adler2019},
which solves the problem of sampling
from a distribution of PDs with a given fixed number of points.

Our RJ-MCMC approach provides two benefits.
Firstly, the larger space of PDs associated with RJ-MCMC
yields samples of PDs that adhere more closely
to topological features present
in a given dataset;
for a more concrete quantification of this argument,
see Section~\ref{sec:comparison}.
Secondly, the number of points in a PD is an unknown hyperparameter
in the presence of noisy data.
RJ-MCMC samples PDs without
conditioning on a specific value of this hyperparameter.
Hence, uncertainty associated with the number of PD points 
is automatically  accounted for by RJ-MCMC.

\section{Methodology}
\label{sec:main}

In this section,
we present our proposed methodology.
More specifically,
we introduce a PIPP model for PDs
based on pairwise interactions of PD points (Section~\ref{sec:model}),
a method for inferring the parameters of the model (Section~\ref{sec:parameter}),
and a RJ-MCMC algorithm to sample PDs
represented by the model (Section~\ref{sec:sampling}).

\subsection{Modeling PDs as PIPP\textbf{s}}
\label{sec:model}

% In this section, we define a PIPP model for PDs. 
In Definition~\ref{def:pd_via_pipp},
we introduce the PIPP representation of a PD.
Subsequently, we specify the spatial density
of PD points and the interaction function
between pairs of PD points in this PIPP representation
of a PD.

\begin{definition}[PIPP representation of a PD]
\label{def:pd_via_pipp}
Let $D = (d_1, \dots, d_n)$ be a PD
in the wedge $\mathbb{W}$ as defined in Equation~\eqref{eq:wedge}.
We assume that the points $(d_1, \dots, d_n)$ of $D$
admit a PIPP on $(\mathbb{W}, \mathcal{W}, \lambda)$
with pdf $f(D \mid \theta )$. According to 
Equations~\eqref{eqn:pairwise_int_density} and ~\eqref{eq:g}, the PIPP density of $D$ is given as follows
\begin{equation}
\label{eqn:pd_via_pipp}
f(D\mid\theta)
% =\frac{1}{Z(\theta)}\prod_{i=1}^ns(x_i)\prod_{i<j}h_{\theta}(x_i,x_j)\\
=\frac{1}{Z(\theta)}
\prod_{i=1}^ns (d_i)
\prod_{i<j} h_{\theta} (d_i,d_j).
\end{equation}
where $\theta\in\mathbb{R}^{k}$ is a parameter vector.
$Z(\theta)$ is the normalizing constant of $f(D\mid\theta)$.
\end{definition}

% In the PIPP representation of a PD $D=(d_1, \dots, d_n)$,
In Equation~\eqref{eqn:pd_via_pipp},
the points $(d_1, \dots, d_n)$ in $D$ admit a spatial pattern function $s$
% which appears in Equation~\eqref{eqn:pairwise_int_density}.
% We hereafter set $s$ to be the spatial density
induced by the Voronoi diagram
$\{T_1,\ldots,T_n\}$,
where $T_i,~i=1,\ldots,n$,
is the Voronoi cell associated with point $d_i$.
The interaction $h_{\theta}(d_i,d_j)$
between two points $d_i$ and $d_j$ of $D$
appears in Equation~\eqref{eqn:pd_via_pipp}.
% in the PIPP representation of a PD $D$. % $D= (d_1, \dots, d_n)$.
We hereafter set $h_{\theta}(d_i,d_j)$ to be
the piece-wise constant pairwise interaction given by
Equation~\eqref{pcpi_function}.

% \vspace{0.1in}
%\begin{itemize}
	%\item \textbf{Spatial pattern term $s$:}
	%\subsubsection{PCPI function and  $h_{\theta}$}
% \noindent	\textbf{Spatial density $s$}.

\subsection{Parameter estimation}
\label{sec:parameter}

The PIPP density value $f(D\mid\theta)$ of a PD $D$
is given by Equation~\eqref{eqn:pd_via_pipp}.
Our goal is to sample PDs from the PIPP density $f(\cdot\mid\theta)$.
In other words, we aim at sampling PDs
that share the same distribution of points with $D$.
We do not have knowledge of parameter $\theta$,
so in this section we provide a way of obtaining
an estimator $\hat{\theta}$ of $\theta$.
Given $\hat{\theta}$,
we construct an RJ-MCMC sampler that
generates PDs from the target PIPP density
$f(\cdot\mid\hat\theta)$ in Section~\ref{sec:sampling}.

To estimate $\theta$, all the available information is
encoded in the PD, $D,$ generated from a given dataset.
Recalling that the points of $D$ admit a PIPP representation,
an approximate estimator $\hat\theta$ of $\theta$ can be
acquired based on the Berman-Turner approach 
according to Equation~\eqref{eqn:bt_logpseudolik}.
More specifically,
$\hat\theta = \argmax_{\theta} \log \tilde{L}(\theta\mid D) $
is computed by maximizing the approximate log-pseudolikelihood
\begin{align*} 
\nonumber&\log \tilde{L}(\theta\mid D) \approx\\&
\sum_{j=1}^m
% \left(
\mathbbm{1}_{\{u_j\in D\}}
\log\mathcal{I}(u_j, D \mid \theta)-
A_j\mathcal{I}(u_j, D\mid \theta),
% \right),
\label{eqn:bt_logpseudolik_d}
\end{align*}
where
$\{u_1,\ldots,u_m\}$ are points in the wedge $\mathbb{W}$
in which $D$ lives, satisfying
$D\subseteq\{u_1,\ldots,u_m\}$.
A Voronoi cell
$T_j,~j=1\ldots,m$, 
of area $A_j$
is associated with point $u_j$,
making up a Voronoi diagram $\{T_1,\ldots,T_m\}$.
The area $\sum_{j=1}^{m}A_j$
is equal to the area of $\mathbb{W}$.
The conditional intensities 
$\mathcal{I}(u_j, D \mid \theta)$
are given by Equation~\eqref{eqn:conditional intensity},
where $s$ is the spatial pattern function induced by
$\{T_1,\ldots,T_m\}$.

The points $\{u_1,\ldots,u_m\}\setminus D$,
which are additional points not in $D$,
constitute a hyperparameter.
If $D$ is a dense PD,
then a practical choice is to set $\{u_1,\ldots,u_m\}\setminus D = \varnothing$
and therefore $\{u_1,\ldots,u_m\}=D$.
If $D$ is a sparse PD with a relatively small number of points,
it is possible to augment it with synthetic points,
in which case $\{u_1,\ldots,u_m\}$ is a strict superset of $D$
containing the original points in $D$
and the synthetic points $\{u_1,\ldots,u_m\}\setminus D$.
Irrespective of how the hyperparameter
$\{u_1,\ldots,u_m\}\setminus D$ is tuned,
it is used only to compute $\hat\theta$.

{{
Subsequently, samples are drawn from the target density
$f(\cdot\mid\hat\theta)$,
as explained in Section~\ref{sec:sampling}
by using the estimated value $\hat\theta$.
In other words,
the PIPP log-pseudolikelihood of Definition~\ref{def:pseudolikelihood}
is used for acquiring the estimate $\hat\theta$ only,
and it is not used in the sampling process.
Having obtained $\hat\theta$,
the PIPP density $f(\cdot\mid\hat\theta)$ of Definition~\ref{def:pd_via_pipp}
becomes the target density from which PD samples are drawn via RJ-MCMC.}}

\subsection{Sampling PDs}
\label{sec:sampling}

The key contribution of this work is an RJ-MCMC algorithm for 
generating random samples of PDs modeled as pairwise 
interaction point processes. 
% The primary goal
% is to generate PDs with topological properties present in a given dataset
% and to conduct statistical inference using the generated PDs.
Notably, our RJ-MCMC algorithm can be utilized to perform inference
based on topological features elicited via PD augmentation,
especially when the sample size of a given dataset is relatively small.
This section starts by
outlining
the three types of moves allowed by RJ-MCMC in the space of PDs
and by providing an informal description of the RJ-MCMC acceptance probabilities
for these moves.
Subsequently, it states formally the RJ-MCMC sampling scheme.

\subsubsection{RJ-MCMC for PDs: outline}
\label{sec:outline}

Our RJ-MCMC sampler explores the PD space via three moves;
(i) a point in a PD can be moved from one location to another
without changing the total number of points,
(ii) a point can be added to a PD, and
(iii) a point can be removed from a PD. 
%The sampler chooses uniformly at random between these three types of moves,
%that is relocation, addition or removal of points.
In particular, the sampling process consists of two types of MCMC updates.
A MWG update relocates points in a PD 
via random-walk Metropolis steps, similar to \cite{Adler2017}.
Furthermore, an RJ-MCMC update adds a point to the PD or removes a point from it,
thus yielding a novel PD sampler.

The  probabilities of
changing the location of a selected point,
of adding a new point,
and of removing a selected 
point
are denoted by
$p_m$, $p_a$, and $p_r$, respectively.
At each RJ-MCMC iteration,
a type of move is chosen randomly according to a categorical distribution
$\text{Categorical}(p_m, p_a, p_r)$
with event probabilities $p_m$, $p_a$, and $p_r$.
% For instance, a move can be chosen uniformly at random,
% that is $p_m=p_a=p_r=1/3$.

Let $D^{(l)} = (d_1^{(l)}, \dots, d_{\lvert D^{(l)}\rvert}^{(l)})$
be the current PD with $\lvert D^{(l)}\rvert$ points at the $l$-th RJ-MCMC iteration.
A type of move is chosen according to $\text{Categorical}(p_m, p_a, p_r)$.
Subsequently, a candidate PD $D^{*}$ is proposed
subject to the type of chosen move.
If it is chosen to relocate the points of $D^{(l)}$,
then
a new location $d_i^*$ is sampled from a proposal density $q$
and the candidate PD is set to $D^* = 
(d_1^{(l)}, \dots, d_{i-1}^{(l)}, {d_i}^*, d_{i+1}^{(l)},\dots,    
d_{\lvert D^{(l)}\rvert}^{(l)})$
for each $i=1,\ldots, \lvert D^{(l)} \rvert$.
If it is chosen to add a new point $d^*$ to $D^{(l)}$,
then $d^*$ is sampled uniformly in the support of
the underlying point process representing $D^{(l)}$
and the candidate PD is set to $D^* = (D^{(l)}, d^*)$.
If it is chosen to remove a point $d^*$ from $D^{(l)}$,
then a point $d^* = d^{(l)}_i\in D^{(l)}$ is chosen randomly
and the candidate PD is set to 
$D^* = D^{(l)} \setminus  d^*$.

Once a candidate PD $D^*$ has been proposed,
it is accepted with probability
$a(D^{(l)}, D^*)$.
If $D^*$ is accepted, then
the PD
at iteration $l+1$ is set to
$D^{(l+1)}=D^*$,
otherwise $D^{(l+1)}=D^{(l)}$.
In the case of point relocation,
$a(D^{(l)}, D^*)$ is a typical Metropolis-Hastings acceptance probability.
In the case of point addition or removal,
$a(D^{(l)}, D^*)$ is a reversible jump acceptance probability
(see Proposition~\ref{prop_acc}).
An RJ-MCMC algorithm for sampling pairwise interacting point processes
is introduced by \cite{Geyer1994}
and is adapted in the present paper to sample PDs.
As part of this adaptation,
Lemma \ref{lemma} is stated
by modifying a corresponding lemma for point processes in \cite{Geyer1994}
to fit the context of sampling PDs,
{{
which are represented by PIPP densities
(see Definition~\ref{def:pipp})}}.

\begin{lemma}\label{lemma}
{{
Let $D$ be a PD with $\lvert D \rvert$ points, which admit 
a PIPP density $f(\cdot \mid \theta)$
given by Equation~\eqref{eqn:pd_via_pipp}}}.
% associated with the pairwise interacting point process
% \eqref{eqn:Gibbs_density}.
Moreover, it is assumed that the number of points of $D$
in a region $R$ has a Poisson distribution with mean $\lambda(R)$.

Let $d^*$ be a point candidate for addition to $D$.
Assume that $d^*$ has distribution
$\frac{\lambda(\cdot)}{\lambda(R)}$.
The acceptance probability for the candidate PD $D^* = (D, d^*)$ is
\begin{equation} \label{lemma:acF_AT_prob_add}
a(D, D^*) = \frac{f(D^* \mid \theta) \lambda(R)}{f(D \mid \theta) (\lvert D \rvert+1)}.
\end{equation}

If $D = \varnothing$ then the PD chain stays at $D$. Otherwise,
let $d^*\in D$ be a point candidate for removal from $D$.
The acceptance probability for the candidate PD $D^* = D \setminus d^*$ is
\begin{equation} \label{lemma:acF_AT_prob_remove}
a(D, D^*) = \frac{f(D^* \mid \theta) (\lvert D \rvert-1) }{f(D \mid \theta) \lambda(R)}.
\end{equation}
\end{lemma}

\subsubsection{RJ-MCMC for PDs: construction}

Proposition \ref{prop_acc} states the acceptance probabilities
for the three types of moves in the proposed RJ-MCMC scheme.
The proof of Proposition \ref{prop_acc} follows from Lemma \ref{lemma}
and is available in Appendix \ref{app_proof}.
As a brief and informal outline of the proof,
the acceptance probability \eqref{accp_relocation}
follows from a Metropolis-Hastings step for point relocation,
while the acceptance probabilities
\eqref{eq:accp_add} and \eqref{eq:accp_removal}
for point addition and point removal
follow from the reversible jump acceptance probabilities
\eqref{lemma:acF_AT_prob_add} and \eqref{lemma:acF_AT_prob_remove}
of Lemma \ref{lemma}, respectively.

\begin{proposition} \label{prop_acc}
	Consider a random PD on the wedge $\W$ modeled by a PIPP density 
	$f(\cdot\mid\widehat{\theta})$ given by Equation \eqref{eqn:pd_via_pipp}.
	%$f(\cdot\mid\widehat{\theta})$ is parameterized by a
	%spatial density $s$
	%and a PCPI function $h_{\widehat{\theta}}$,
	%conditional on the parameter estimate
	%$\widehat{\theta}$ obtained by maximizing the pseudolikelihood function
	%given by Equation\eqref{eqn: approx_pseudo}.
	The number of points of a PD in a region $\mathbb{W}$ has a Poisson distribution with mean $\lambda(\mathbb{W})$.
	%The PCIP function $h_{\widehat{\theta}}$ is given by Equation \eqref{pcpi_function}.
	Let $D^{(l)} = 
	(d_1^{(l)}, \dots, d_{\lvert D^{(l)} \rvert }^{(l)})$ be the PD at the $l$-th MCMC iteration.
	The acceptance probabilities for generating random PDs from $f(\cdot\mid\widehat{\theta})$
	by relocating, adding or removing points follow.

    Let $D^* = 
		(d_1^{(l)}, \dots, d_{i-1}^{(l)}, {d_i}^*, d_{i+1}^{(l)},\dots,    
		d_{\lvert D^{(l)}\rvert})$ 
	    be the candidate PD for the relocation move,
	    where $d_i^*$ is chosen according to a proposal density $q$. The 
		acceptance probability for $D^*$ is 
%		\vspace{-0.2in}
		\begin{align}
		&\nonumber a(D^{(l)},D^*) =\\& \min\left\{1, \frac{s ({d_i}^*)  
		g({D^*\mid\widehat{\theta}}) q({d_i}^{(l)})}{s ({d_i}^{(l)})  
		g({D^{(l)}\mid\widehat{\theta}})q({d_i}^{*})} \right\},\label{accp_relocation}
		\end{align}

    For the addition of a point $d^*$ to $D^{(l)}$,
    we choose $d^*$ uniformly at random in $\W$
	and obtain the candidate PD $D^* = (D^{(l)}, d^* )$. The acceptance probability for $D^*$ is
		\begin{align}   
	 &\nonumber a(D^{(l)},D^*)  =\\&  \min\left\{1, \frac{
	 %\exp \left(\sum_{i=1}^{\vert D^{(l)}\vert} \ln(h_{\widehat{\theta}} (d^{(l)}_i,d^*))\right)
	 \left[\prod_{i=1}^{\vert D^{(l)}\vert} h_{\widehat{\theta}} (d^{(l)}_i,d^*)\right]
		s (d_i^{*}) 
		\lambda{(\mathbb{W})}}{\vert D^{(l)}\vert+1} \right\}. \label{eq:accp_add}
		\end{align}
		
    For the removal of a point $d_i^{(l)}$ from $D^{(l)}$,
		we  choose uniformly at random $d_i^{(l)}$ 
		and obtain the candidate PD $D^* = D^{(l)} \setminus  d^{(l)}_i $.
		The acceptance probability for $D^*$ is
		\begin{align}  
		&\nonumber a(D^{(l)},D^*) =\\& \min\left\{1, \frac{\vert D^{(l)}\vert-1} {
		%\exp \left(\sum_{j  \neq i} \ln(h_{\widehat{\theta}} 		(d^{(l)}_j,d_i^{(l)}))\right)
		\left[\prod_{j\neq i}h_{\widehat{\theta}}	(d^{(l)}_j,d_i^{(l)})\right]
		s (d_i^{(l)}) 
		\lambda{(\mathbb{W})}}\right\}. \label{eq:accp_removal}
		\end{align}
\end{proposition}

The pseudocode of the RJ-MCMC sampler
for generating PDs is summarized by Algorithm \ref{alg}.
% Algorithm \ref{alg} includes three possible moves of PD points,
% namely relocation, addition and removal of points.
% Appendix \ref{app_design} provides two alternative scenarios;
% MWG sampling of PDs with fixed number of points
% allowing only relocation of points,
% and RJ-MCMC sampling of PDs allowing only addition and removal of points.
Section~\ref{sec:comparison}
% demonstrates the utility of Algorithm \ref{alg} via examples.
% Appendix \ref{app_design}
provides an experimental validation of the
relative advantages of Algorithm \ref{alg} in comparison to
% two alternative MCMC schemes,
% namely
% % in comparison to
% (i) a special case of our RJ-MCMC sampler that allows only addition and removal of PD points
% and (ii)
a MWG sampler of PDs with fixed number of points \cite{Adler2019}.
% % other two considered MCMC schemes for sampling PDs.

\begin{algorithm}[!ht] 
	\caption{RJ-MCMC sampling of PDs}
	\label{alg}
	\begin{algorithmic}[1]
		\State{\textbf{Input}:
		initial PD $D^{(0)} = (d_1^{(0)}, \dots, d_{\vert D^0 \vert}^{(0)})$
		}
		\State{\textbf{Input}:
		probabilities $(p_m, p_a, p_r)$
		}
		%\State{Choose an initial persistence diagram $D^{(l)} = (d_1^{(l)}, \dots, 
		%d_{|D^l|}^{(l)})$ randomly}
		\State{}
		
		\For{$l\in\{1,...,N\}$}
		%\State{Choose a number from $\{1,2,3\}$ randomly and call it $\gamma$}
		%\State {\textbf{Input}: Move probabilities $p_a$ -- to add a point, $p_r$ 
		%-- to remove a point, and $p_m$ -- to move a point such that $p_a+p_r+p_m = 
		%1$.}
		\State{Sample $\gamma$ from
		$\mbox{Categorical}(p_m, p_a, p_r)$}
		% $\gamma\in\{1,2,3\}$ from
		% $\gamma \sim \mbox{Categorical}(p_m, p_a, p_r)$
        \State{}

		\If{$\gamma =1$}
		  \State{Choose $i$ randomly from 1 to $\vert D^{(l)}\vert$}
		    \State{Sample $d_i^*$ from proposal density $q$}
	        %\State{$D^* = (d_1^{(l)}, \dots, d_{i-1}^{(l)}, d_i^*, 
			%  d_{i+1}^{(l)}\dots,    d_{\vert D^l \vert}^{(l)})$}
			\State{$D^* = (d_1^{(l)}, \dots, d_i^*, \ldots, d_{\vert D^{(l)} \vert}^{(l)})$}
		    \State{Compute $a(D^{(l)},D^*)$ from Eq
		      \eqref{accp_relocation}}
		    \State{Sample $u$ from uniform $\mathcal{U}(0,1)$}
		    \If{$u < a(D^{(l)},D^*)$}
		      \State{$D^{(l+1)} = D^*$}
		    \Else{}
		    \State{$D^{(l+1)} = D^{(l)}$}
		    \EndIf
		%\State{}
		
		\ElsIf {$\gamma =2$}
		\State{Sample $d^*$ uniformly at random in $\W$}
		\State{$D^* = (D^{(l)},  d^*)$}
		\State{Compute  $a(D^{(l)},D^*)$ from Eq \eqref{eq:accp_add}}
		\State{Sample $u$ from uniform $\mathcal{U}(0,1)$}
		\If{$u < a(D^{(l)},D^*)$}
		\State{$D^{(l+1)} = D^*$}
		% \State{$\vert D^{(l+1)}\vert = \vert D^{(l)} \vert + 1$}
		\Else{}
		\State{$D^{(l+1)} = D^{(l)}$}
		\EndIf
		%\State{}
		
		\Else{}
		\State{Sample a point $d^*_i$ from $D^{(l)}$}
		\State{$D^* = D^{(l)}
		\setminus   d^{(l)}_i $}
		\State{Compute  $a(D^{(l)},D^*)$ from Eq \eqref{eq:accp_removal}}
		\State{Sample $u$ from uniform $\mathcal{U}(0,1)$}
		\If{$u < a(D^{(l)},D^*)$}
		\State{$D^{(l+1)} = D^*$}
		% \State{$ \vert D^{(l+1)}\vert = \vert D^{(l)} \vert -1$}
	    \Else{}
		\State{$D^{(l+1)} = D^{(l)}$}
		\EndIf		
		\EndIf
        \EndFor
	\end{algorithmic}
\end{algorithm}

%\noindent \textbf{MCMC Diagnostics:} \textcolor{red}{Theo will add this part}
% \begin{figure}[h!]
% 			\centering
% 				\subfloat[]{	
%\includegraphics[width=1.2in,height=1.2in]{Figures/adler_2_circles.png}} 
%\hspace{0.2in}
% 			\subfloat[]{	
%\includegraphics[width=1.5in,height=1.2in]{Figures/adler_2_circles_pd.png}}\hspace{0.2in}
% 			\subfloat[]{	
%\includegraphics[width=1.2in,height=1.2in]{Figures/adler_2_circles_tilted_pd.png}}
% 				\vspace{-0.1in}
% 			\caption{The data points and the corresponding PD used in Example 
%\ref{sec:sampling example_1} \label{fig:two_circles}} 
% 				\vspace{-0.1in}
% 		\end{figure}

% Acknowledgements should go at the end, before appendices and references
\begin{figure}[!ht]
	\centering
	\subfloat[]{	
	\includegraphics[width=0.21\textwidth]{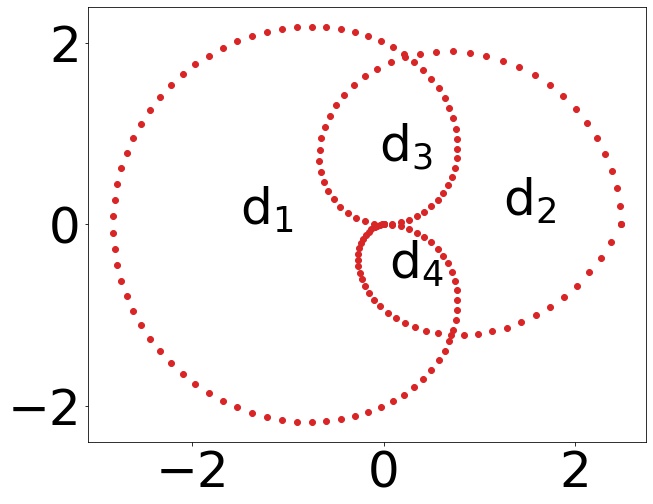}\label{fig:polar_curve_pds_a}}\hspace{0.1in}
	\subfloat[]{	
	\includegraphics[width=0.21\textwidth]{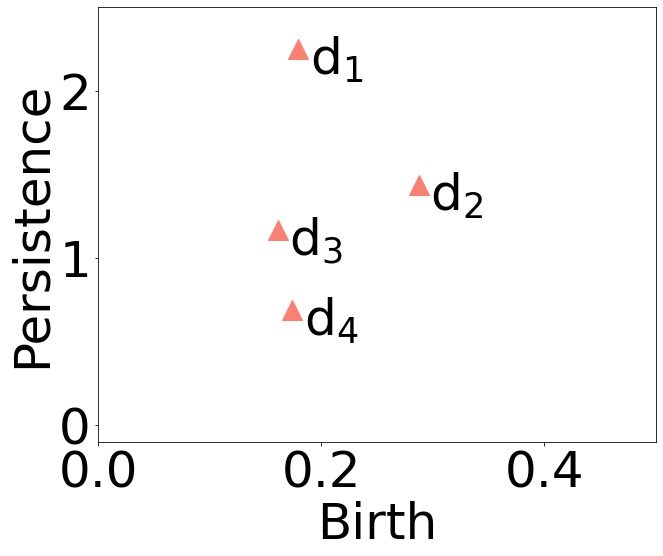}\label{fig:polar_curve_pds_b}}\hspace{0.1in}
	\subfloat[]{	
	\includegraphics[width=0.21\textwidth]{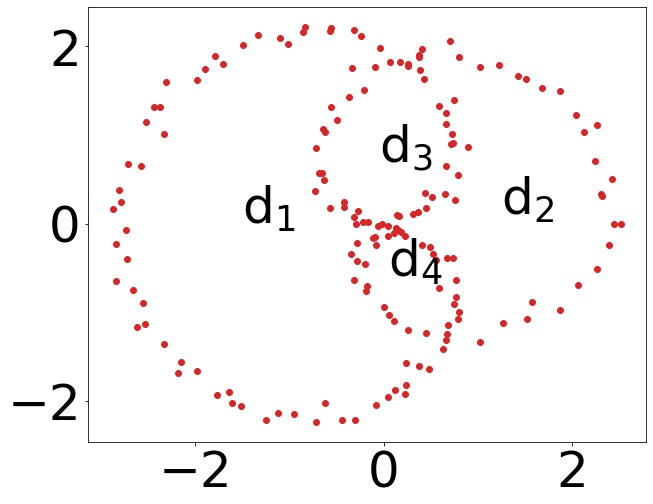}\label{fig:polar_curve_pds_c}}\hspace{0.1in}
	\subfloat[]{	
	\includegraphics[width=0.21\textwidth]{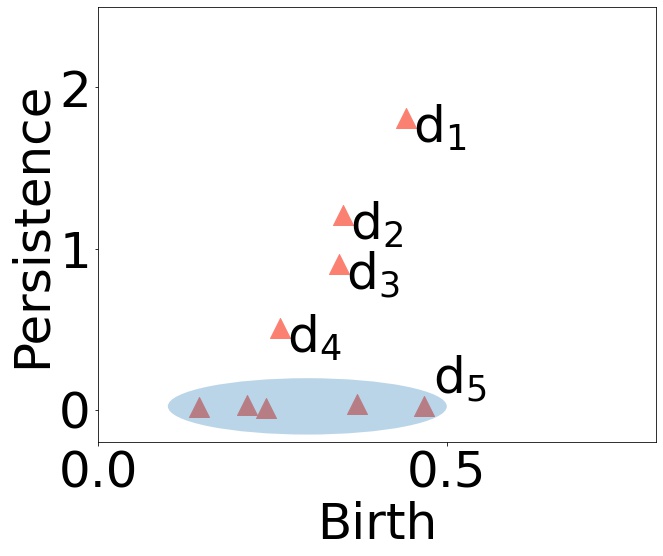}\label{fig:polar_curve_pds_d}}\hspace{0.1in}
	\subfloat[]{	
	\includegraphics[width=0.21\textwidth]{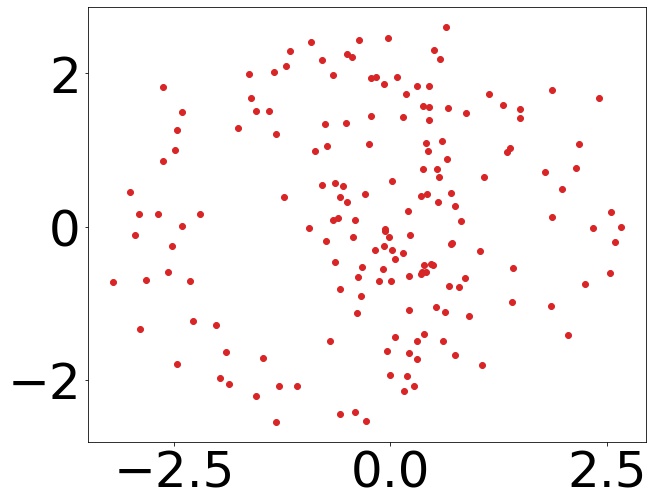}\label{fig:high_noise_cloud}}\hspace{0.1in}
	\subfloat[]{	
	\includegraphics[width=0.21\textwidth]{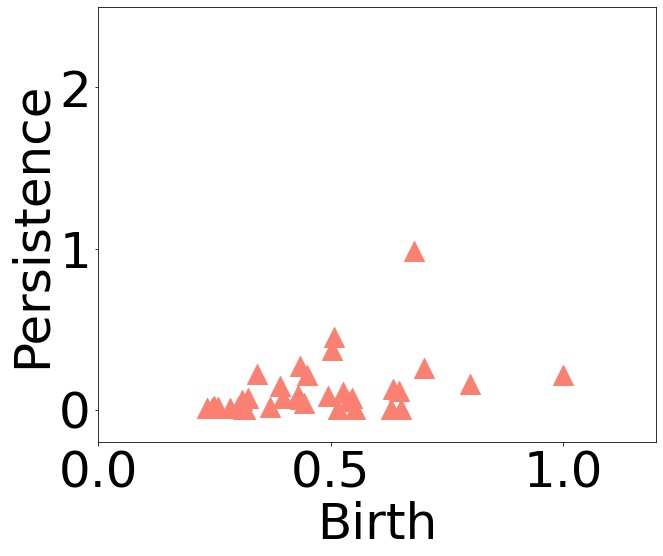}\label{fig:high_noise_pd}}\hspace{0.1in}
	\caption{(a): a random sample of $170$ points from the asymmetric knot
	defined by Equation~\eqref{eq:as_knot}.
	(c) and (e): noisy versions of the random sample in (a), after adding noise to each point
	in (a), with noise having been drawn
	from a normal distribution centered at the point with variance
	$\sigma^2=0.005$ and $\sigma^2=0.1$, respectively.
	(a), (c) and (e) are displayed in Cartesian coordinates.
	(b), (d) and (f) are the PDs of one-dimensional features
	extracted from the simulated datasets in (a), (c) and (e),
	respectively. The PDs have been generated using Vietoris-Rips filtration, as discussed in Section \ref{subsec:vr}.}
% 			(c) is a noisy version of  (a), consequently, their persistence 
% 			diagrams show different patterns of interaction between points.
	\label{fig:polar_curve_pds}
\end{figure}

\section{Asymmetric knot example}
\label{sec:motivating example}

% Motivated by the specific spatial settings held by persistence diagrams, we 
%focus on the pairwise interacting point process to capture the underlying 
%spatial structure of them. 
% Let us discuss a motivating example as a justification for applying this 
%point process. 
In this section, we consider two noisy point cloud datasets shown in
Figures \ref{fig:polar_curve_pds_c} and \ref{fig:high_noise_cloud},
which have been generated by adding normally distributed noise
with respective variance
$\sigma^2=0.005$ and $\sigma^2=0.1$,
to the point cloud data of Figure \ref{fig:polar_curve_pds_a}.
The data of Figure \ref{fig:polar_curve_pds_a}
have been generated from the asymmetric knot
\begin{equation}
\label{eq:as_knot}
    \text{knot}(\phi) = 
    \begin{cases}
    \frac{7\sqrt{2}}{4}\cos\left(\frac{\phi}{2}\right), & 0\le\phi<\pi,\\
    2\sqrt{2}\cos\left(\frac{\phi}{2}\right), & \pi\le\phi<3\pi,\\
    \frac{7\sqrt{2}}{4}\left(\frac{\phi-3\pi}{\pi}\right)^{7/5}, & 3\pi\le\phi<4\pi.
    \end{cases}
\end{equation}
%which is a type of polar curve.
%Celtic knots have been observed at electron
%micrographs of closed DNA.

% We generate noisy versions of
% the polar curve of Figure~\ref{fig:polar_curve_pds_a},
% by adding normally distributed noise with
% two levels of variance:
% low ($\sigma^2 = 0.005$),
% and high ($\sigma^2 = 0.1$);

Figures \ref{fig:polar_curve_pds_b},
\ref{fig:polar_curve_pds_d} and
\ref{fig:high_noise_pd}
show the respective PDs of the
noiseless point cloud (Figure \ref{fig:polar_curve_pds_a}),
of the point cloud with low level of noise 
($\sigma^2 = 0.005$, Figure \ref{fig:polar_curve_pds_c})
and of the point cloud with high level of noise
($\sigma^2 = 0.1$, Figure \ref{fig:high_noise_cloud}).
These PDs have been generated using Vietoris-Rips filtration (see Section \ref{subsec:vr}).
% and have found that these can topologically be knotted circles.
We focus on $1$-dimensional holes in the PDs
shown in Figures \ref{fig:polar_curve_pds_b},
\ref{fig:polar_curve_pds_d} and \ref{fig:high_noise_pd},
as such holes characterize the prominent shape features of the asymmetric knot.  
% The PDs suggest that the spatial locations  exhibit regularity in their 
%spacing, in other words,  there is a possible interaction between points. 

Section \ref{sec:sampling example_3} deploys our RPDG algorithm
to sample PDs from the PIPP density of the PD of Figure \ref{fig:polar_curve_pds_d}.
Section \ref{sec:comparison} compares RPDG
with an existing PD sampling method \cite{Adler2019}; for this comparison,
we consider the point cloud data
of Figure \ref{fig:polar_curve_pds_c}
and the corresponding PD of Figure \ref{fig:polar_curve_pds_d}.

\subsection{RPDG illustration}	
\label{sec:sampling example_3}

% First, we discuss how we implement the RPDG technique to generate 1000 samples 
% of the PD  in Fig. \ref{fig:polar_curve_pds} (d).
% of  the noisy version of the polar curve Fig. (\ref{fig:polar_curve_pds} (b)) generated by VR filtration. 

We illustrate how RPDG can be used to sample PDs
from the target PIPP density of the PD of Figure~\ref{fig:polar_curve_pds_d}.
% Recall that the PD of Figure~\ref{fig:polar_curve_pds_d}
% has been generated from the point cloud data
% of Figure~\ref{fig:polar_curve_pds_c}.
% The point cloud data of Figure~\ref{fig:polar_curve_pds_c}
% have been generated by adding noise to points
% that have been simulated from the asymmetric knot of Equation~\eqref{eq:as_knot}.
In particular,
Section \ref{subsec:sampling}
provides an example of how to setup RJ-MCMC sampling of PDs from the target PIPP density,
% related to the asymmetric knot of Figure~\ref{fig:polar_curve_pds_c},
while Section~\ref{sec:rpdg_dist}
introduces a notion of running average distance in the space of PDs
to assess quality of PD sampling from a topological point of view.
Section \ref{subsec:sens} presents some sensitivity analysis
for the employed RJ-MCMC sampling scheme
under different levels of noise in the original point cloud data from
which the PD of Figure~\ref{fig:polar_curve_pds_d} has been generated.

\subsubsection{RJ-MCMC sampling}
\label{subsec:sampling}

The point cloud data in Figure \ref{fig:polar_curve_pds_c} consist of four loops,
each of different size.
The corresponding PD in Figure \ref{fig:polar_curve_pds_d} has four points
$d_i,~i=1,2,3,4$, associated with the loops of Figure \ref{fig:polar_curve_pds_c}.
Due to noise in the data of Figure \ref{fig:polar_curve_pds_c},
several other PD points with lower persistence values are spawn,
visualized inside the blue oval of Figure \ref{fig:polar_curve_pds_d}.
These PD points inside the blue oval
do not correspond to any of the four topological features (loops),
they are considered to be noise,
and they are thus clustered together.

To sample PDs via RJ-MCMC as outlined by Algorithm \ref{alg},
it is required to setup four components.
More specifically,
we specify a proposal density $q$ for sampling PD points,
the target PIPP density $f(\cdot \mid \hat\theta)$,
the initial PD $D^{(0)}$
and probabilities $(p_m, p_a, p_r)$.
% Such specification details can be found
% in this section and in Appendix X.

We set the pertinent mixture
%\[
$
q(\cdot) = 
\sum_{i=1}^{5}w_i\mathcal{N}^*(\cdot \mid\mu_i, \sigma_i^2\,I)
$
%\]
as proposal density to sample a candidate PD point from the wedge $\mathbb{W}$
over which the PD of Figure \ref{fig:polar_curve_pds_d} is defined.
% \begin{equation*}
% \label{eqn:gm}
% \end{equation*}
$\mathcal{N}^{*}$ is a bivariate truncated normal density supported on $\W$,
$I$ is the identity matrix,
$(\mu_1,\mu_2, \mu_3,\mu_4,\mu_5)$ are the mixture component means,
$(\sigma_1^2,\sigma_2^2,\sigma_3^2,\sigma_4^2,\sigma_5^2)$ are the mixture component variances,
and $(w_1,w_2,w_3,w_4, w_5)$ are the mixture component weights.
Table~\ref{tab:mixture_comp_values}
in Appendix~\ref{app_knot}
shows the values of $\mu_i,\sigma_i^2$ and $w_i$.
This proposal mixture has been chosen empirically to capture the shape behavior of Figure \ref{fig:polar_curve_pds_d}, as topologically expressed in the associated PD (Figure \ref{fig:polar_curve_pds_d});
notice that the mixture component means $\mu_i,~i=1,2,3,4$,
of Table~\ref{tab:mixture_comp_values}
are placed on the PD points $d_i$ of Figure \ref{fig:polar_curve_pds_d},
while the fifth mean $\mu_5$ is placed on the blue oval of Figure \ref{fig:polar_curve_pds_d}.
Appendix~\ref{app_knot} provides the hyperparameters,
whose values have been empirically set,
for the target PIPP density $f(\cdot \mid \hat\theta)$
% based on the asymmetric knot
of this section.
% We compute the coefficient estimate $\widehat\theta$
% via the process described in Section~\ref{sec:parameter}.
We initialize Algorithm~\ref{alg}
by setting the PD of Figure~\ref{fig:polar_curve_pds_d}
as the initial PD $D^{(0)}$
and by setting $p_a = p_r = p_m = 1/3$.
We then
% use $\widehat\theta$ to
generate $N = 100,000$ samples of PDs
via Algorithm \ref{alg}.

% The proposal distribution $q$ is considered to be the restricted Gaussian mixture density of Equation \eqref{eqn:gm}.  One of the central motivations to estimate samples of PDs is to differentiate noise from topological 
% signals. In order to compare with the Metropolis-within-Gibbs (MWG) algorithm in \cite{Adler2019}, we generate statistical quantities from the samples of PDs following the method showed in \cite{Adler2019}. 

\subsubsection{Running average distance}
\label{sec:rpdg_dist}

We introduce
an empirical metric to assess the capacity of RPDG
% (Algorithm~\ref{alg})
to sample PDs that preserve topological structure.
To this end,
we propose the running average distance
between the persistence order statistics of the noiseless PD
and the persistence order statistics of PDs generated via RPDG.
Typically, the noiseless PD is not known given a dataset.
However, in the controlled experimental setup of this RPDG illustration,
we know the ground truth of noiseless PD (see Figure~\ref{fig:polar_curve_pds_b}).

Let $d^{\text{true},{(i)}}$ be the $i$-th largest persistence value
of the points in the noiseless PD of Figure~\ref{fig:polar_curve_pds_b}.
Moreover, let $d^{\text{sim}, (i)}_k$
be the $i$-th largest persistence value of the points
in the $k$-th PD sample of a realized chain of PDs, where $k = 1,\ldots N$.
We define the running average distance
for the $i$-th largest persistence value
up to the $n$-th PD sample to be
\begin{equation}\label{eq:dist}
% $
\mbox{dist}_{n}(i) =
\frac{1}{n}\sum_{k=1}^n
\vert d^{\text{sim}, (i)}_k-d^{\text{true},{(i)}}  \vert ,
% $
\end{equation}
where $n = 1,\ldots, N$. {{This distance has been chosen to examine the closeness of topologically prominent points in the simulated PDs to the four prominent points in the original PD.}}
In Sections
\ref{subsec:sens}
and
\ref{sec:comparison},
we generate a trace plot of
running average distance $\mbox{dist}_{n}(i)$ 
against RPDG (or MWG) iteration $n$ for each $i\in\{1,2,3,4\}$,
since the noiseless PD of Figure~\ref{fig:polar_curve_pds_b}
has four topologically prominent persistence values $d_i,~i=1,2,3,4$.

The notion of running average distance can be applied to a real dataset
by replacing the noiseless PD with the PD generated from a dataset.
In such a case, the running average distance would quantify
the distance of RPDG samples from the PD of the dataset.

\subsubsection{Sensitivity analysis}
\label{subsec:sens}
\begin{figure}
    \centering
    \includegraphics[width=0.47\textwidth]{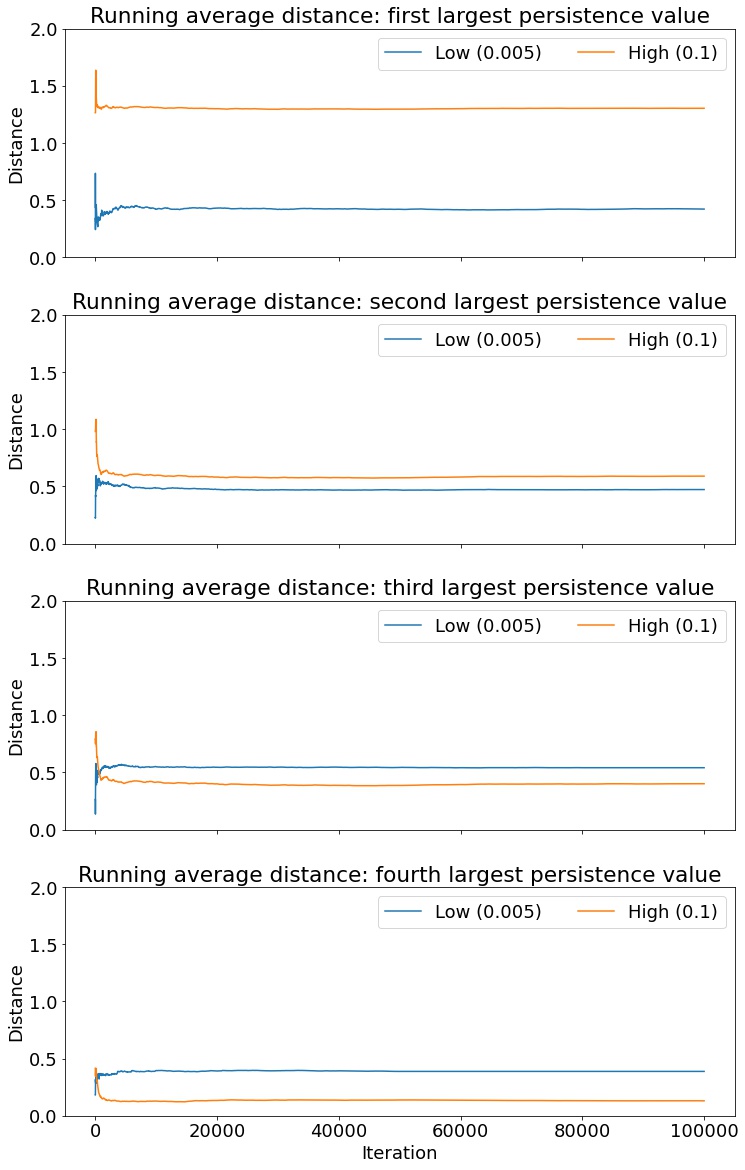}
    \caption{Sensitivity analysis for RPDG.
    Each plot displays three lines:
    a blue and orange line
    representing the running average distance for a persistence value
    using RPDG chains initialized at PDs associated with
    low ($\sigma^2=0.005$)
    and high ($\sigma^2=0.1$)
    levels of noise, respectively.}
    \label{fig:sensitivity}
\end{figure}

Recall the two scenarios of point clouds with low and high noise given in Figures \ref{fig:polar_curve_pds_c}  and \ref{fig:high_noise_cloud}, respectively, as well as their corresponding persistent diagrams in Figures \ref{fig:polar_curve_pds_d} and \ref{fig:high_noise_pd}.
% We generate noisy versions of
% the polar curve of Figure~\ref{fig:polar_curve_pds_a},
% by adding normally distributed noise with
% two levels of variance:
% low ($\sigma^2 = 0.005$),
% and high ($\sigma^2 = 0.1$);
% see Figures~\ref{fig:polar_curve_pds_c} and~\ref{fig:polar_curve_pds_d}
% for the plots related to low level noise, and Figures~\ref{fig:high_noise_cloud} and~\ref{fig:high_noise_pd} for the ones associated to high noise.
For each noise level,
we sample $100,000$ PDs via RPDG.
Subsequently, we compute the running average distance
given by Equation \eqref{eq:dist} for each of the four largest persistence values,
as explained in Section~\ref{sec:rpdg_dist},
and display these distances in Figure~\ref{fig:sensitivity}.

In each of the four plots, RPDG converges in the sense that the distance of Equation \eqref{eq:dist} converges.
As one may expect, the high noise example of Figure~\ref{fig:high_noise_pd} produces the largest distance in the first and second largest persistence value
in comparison to the low noise example of Figure~\ref{fig:polar_curve_pds_d}.
In contrast, for the case of the third and fourth largest persistence value, the running average distance of the low noise case is larger than the one of the high noise case.
This is not unexpected since the associated PD, depicted in Figure \ref{fig:high_noise_pd}, contains the majority of its points in the region of the underlying true PD points $d_3$ and $d_4$ (small loops); see Figure  \ref{fig:polar_curve_pds_b}. Thus, a large number of samples are drawn from that area, which in turn leads to a small deviation from the underlying ground truth.

% \begin{figure}[!ht]
% 	\centering
% 	\subfloat[]{	
% 	\includegraphics[width=0.21\textwidth]{Figures/AsymExampleMidNoise.jpg}\label{fig:mid_noise_cloud}}\hspace{0.1in}
% 		\subfloat[]{	
% 	\includegraphics[width=0.21\textwidth]{Figures/AsymExampleMidNoisePD.jpg}\label{fig:mid_noise_pd}}\hspace{0.1in}
% 	\subfloat[]{	
% 	\includegraphics[width=0.21\textwidth]{Figures/AsymExampleHighNoise.jpg}\label{fig:high_noise_cloud}}\hspace{0.1in}
% 	\subfloat[]{	
% 	\includegraphics[width=0.21\textwidth]{Figures/AsymExampleHighNoisePD.jpg}\label{fig:high_noise_pd}}\hspace{0.1in}
% 	\caption{(a): Noisy version
% 	of the point cloud data of Figure~\ref{fig:polar_curve_pds_a},
% 	generated by adding normally distributed noise with medium noise level ($\sigma^2 = 0.01$)
% 	at each point in Figure \ref{fig:polar_curve_pds_a}.
% 	(c): Noisy samples generated similarly to (a), but with high noise level ($\sigma^2 = 0.1$).
% 	(b) and (d) are the PDs of one-dimensional features extracted from the simulated datasets in (a) and (c), respectively.}
% 	\label{fig:noisysets}
% \end{figure}

\subsection{Comparison with MWG}
\label{sec:comparison}

In this section, we compare RPDG with the MWG sampling scheme proposed in \cite{Adler2019}. 
{{
MWG has a set of hyperparameters, which have been empirically tuned.
Subsequently, the parameters involved in MWG sampling are
estimated in accordance with~\cite{Adler2019}.}}

We sample $100,000$ PDs using MWG under the low level noise scenario,
as the intention is to compare the `topological fidelities' of RPDG and MWG
for high quality data,
that is for data relatively clean from noise
admitting an underlying topological structure.
% By using the  MWG sampling scheme we realize $100,000$ samples 
% of PDs. The parameters used for this scheme were estimated in accordance with~\cite{Adler2019}.
We then compute the running average distance
for each of the four largest persistence values
based on MWG PD samples
and on the noiseless PD of Figure~\ref{fig:polar_curve_pds_b},
as described in Section \ref{sec:rpdg_dist}.

\begin{figure}
\centering
\includegraphics[width=0.47\textwidth]{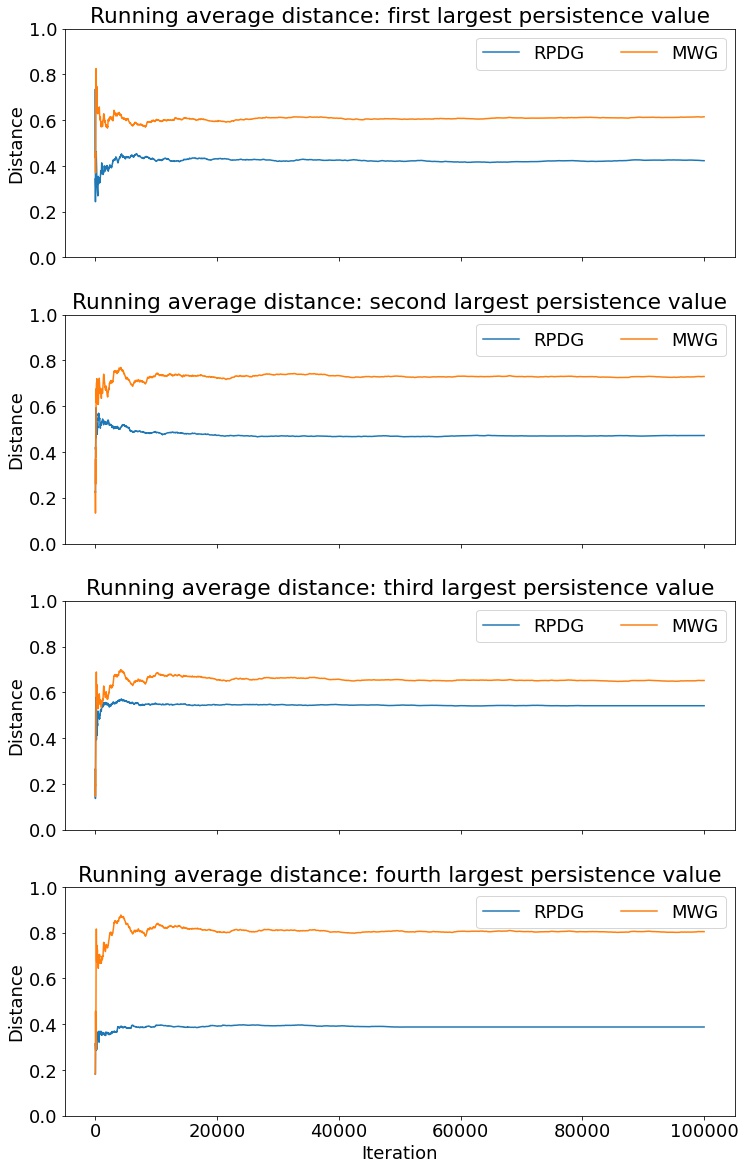}
\caption{A comparison between RPDG and MWG.
Each plot displays two lines:
a blue line
representing the running average distance
for a persistence value based on
RPDG samples and the noiseless PD of Figure~\ref{fig:polar_curve_pds_b},
and an orange line
representing the running average distance
for the same persistence value based on
MWG samples and the noiseless PD of Figure~\ref{fig:polar_curve_pds_b}.}
\label{fig:adler_comparison}
\end{figure}

Figure~\ref{fig:adler_comparison} overlays
the running average distance associated with RPDG and with MWG
for each persistence value.
The displayed running average distances are computed from
one RPDG and one MWG chain realization,
with both chains having the PD of Figure~\ref{fig:polar_curve_pds_d}
as their initial state.
While both sampling methods converge,
RDPG enjoys lower distance from the ground truth in comparison to MWG.

\section{Materials science example}
\label{sec:application}
In this section, we consider a real materials science
dataset collected in an experimental facility.
Section \ref{ssubsec:intro_materials} sets the stage
by introducing the underlying materials science problem,
while Section \ref{subsect:data} describes
the experimental data under consideration herein.
Next a classical Kolmogorov-Smirnov test is considered
in Section \ref{subsec:vanilla_KS} to attack the problem,
while a Kolmogorov-Smirnov test based on RPDG is proposed
in Section \ref{subsec:KS_RPDG}.
The latter approach (as opposed to the former one)
solves the materials science problem,
matching experimental knowledge.

\subsection{Introduction} \label{ssubsec:intro_materials}

Advancement towards high strength steels is of interest in the materials science community. For example, austenitic stainless steels (AuSS) are widely used in various fields from biomedical engineering to automobile industry,  and to everyday life, e.g. see \cite {Na2017} and references therein. Recently synthesized nano-grained (NG) structured AuSS have properties such as superior tensile strength, fatigue strength, and fracture toughness.
Due to their properties, NG AuSS are used as biomaterials to replace structural components of the human body \cite{Murphy2016}. 
% Recently, nano/ultrafine-grained (NG/UFG) structured AuSS have been synthesized with a combination of high strength and good ductility through phase reversion via severe cold reduction followed by annealing \cite{MISRA2015}.
Quantitative microstructure analysis is an important step towards understanding
the structure and behavior of NG AuSS materials.
Electron backscatter diffraction (EBSD) is an experiment that generates data essential for quantitative microstructural analysis.
In particular, it provides 
grain sizes, the morphology of individual grains, crystallographic relationships between phases, and the  Schmid factor.

% of EBSD, which evaluates the ease of dislocation movement within a grain, is useful to relate material structure and properties \cite{Li2019}. 

\begin{figure}[!ht]
	\centering
	\subfloat[]{	
	\includegraphics[width=0.2\textwidth]{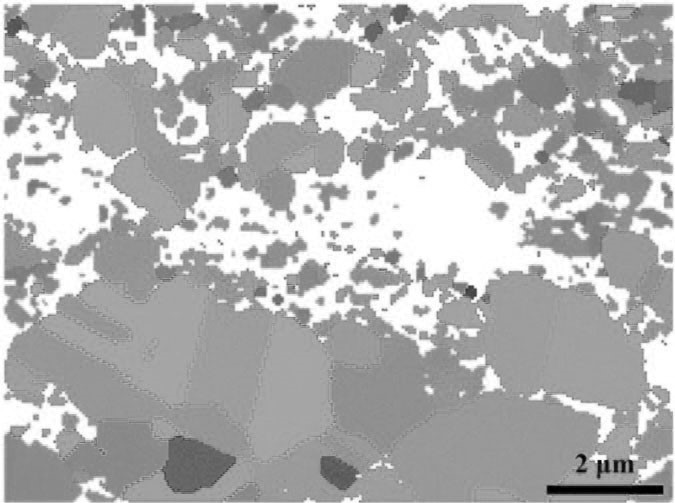}}\hspace{0.1in}
	\subfloat[]{	
	\includegraphics[width=0.2\textwidth]{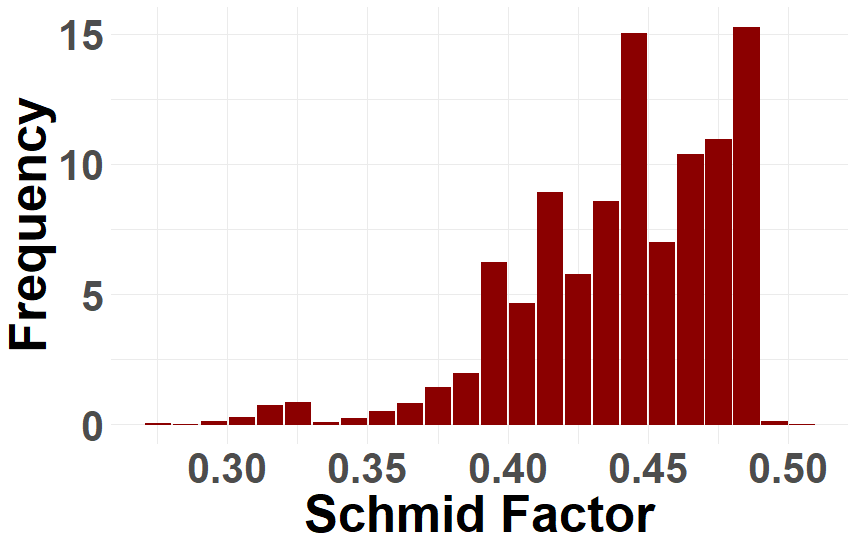}}\hspace{0.1in}
		\subfloat[]{	
	\includegraphics[width=0.2\textwidth]{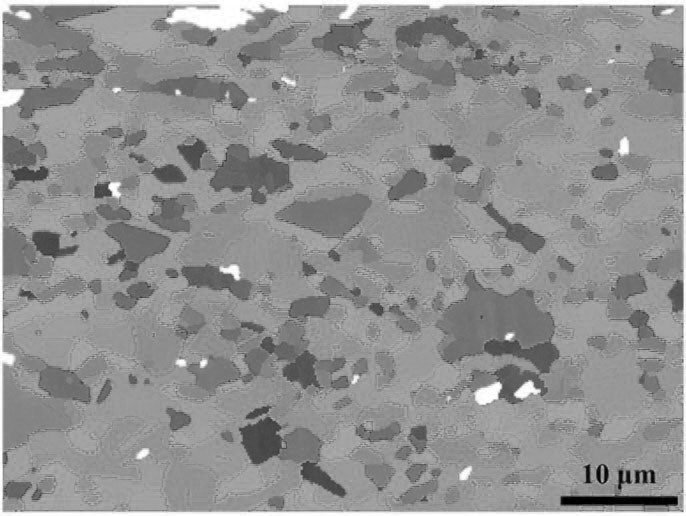}}\hspace{0.1in}
	% \hspace{0.02in}
	\subfloat[]{	
	\includegraphics[width=0.2\textwidth]{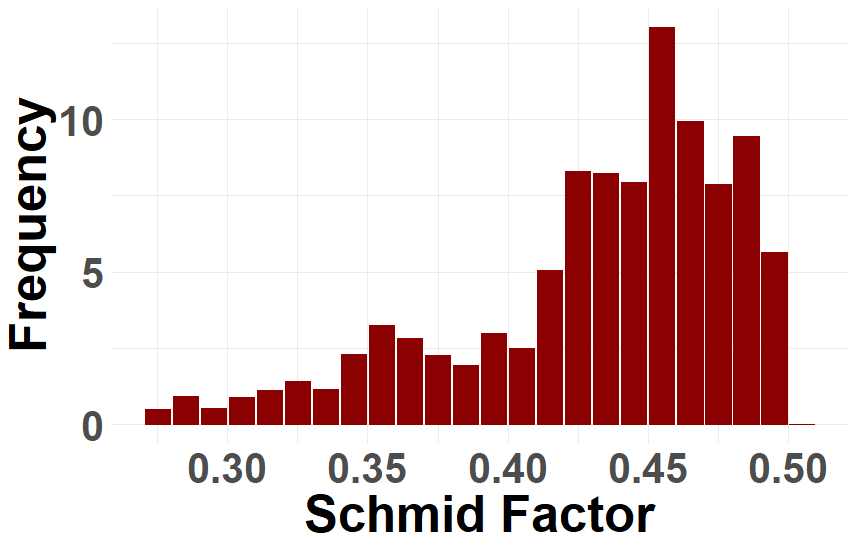}}\hspace{0.1in}\\
		% \hspace{0.02in}
		\subfloat[]{	
	\includegraphics[width=0.2\textwidth]{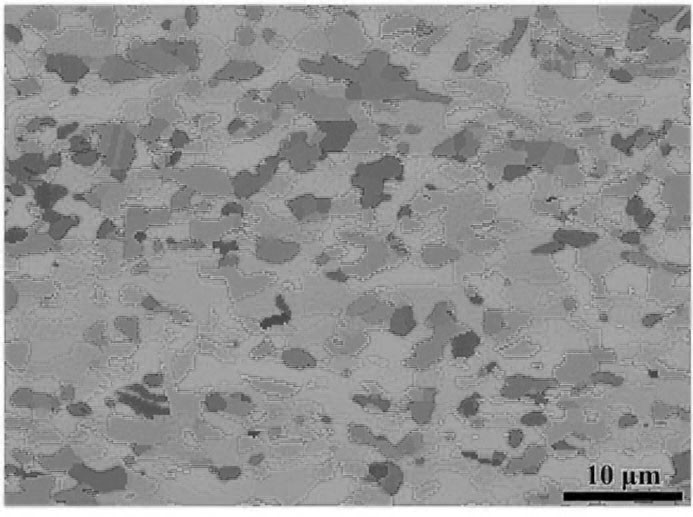}}\hspace{0.1in}
		% \hspace{0.02in}
	\subfloat[]{	
	\includegraphics[width=0.2\textwidth]{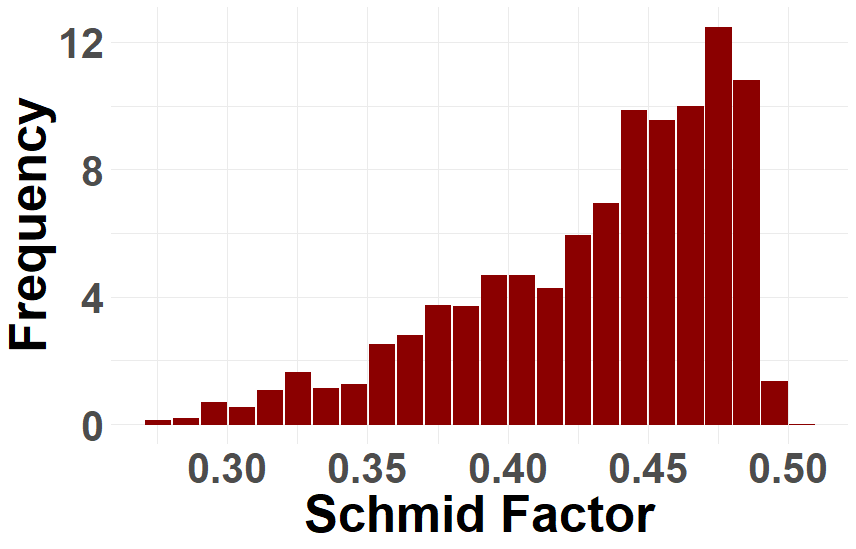}}\hspace{0.1in}
		\subfloat[]{	
	\includegraphics[width=0.2\textwidth]{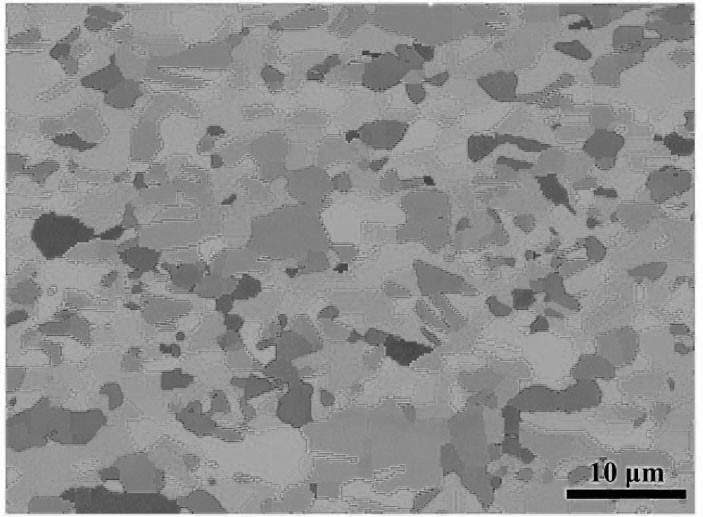}}\hspace{0.1in}
	\subfloat[]{	
	\includegraphics[width=0.2\textwidth]{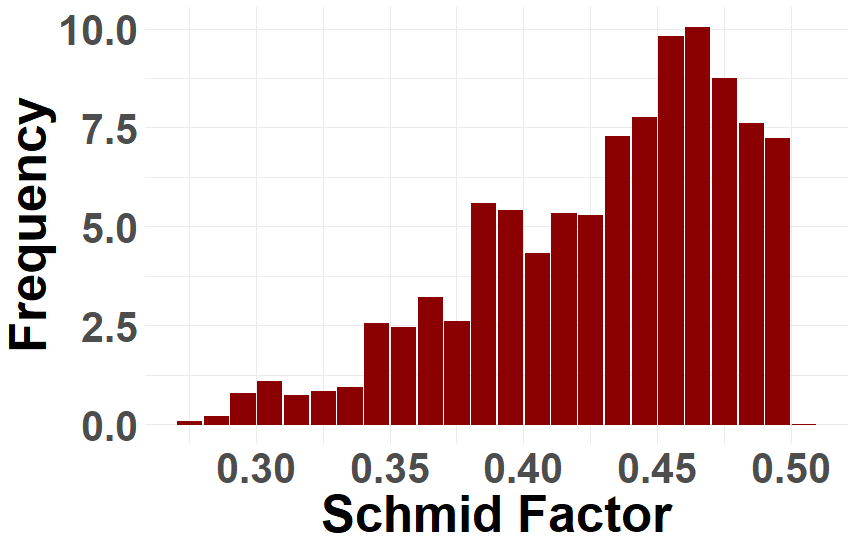}}\hspace{0.1in}
		\subfloat[]{	
	\includegraphics[width=0.2\textwidth]{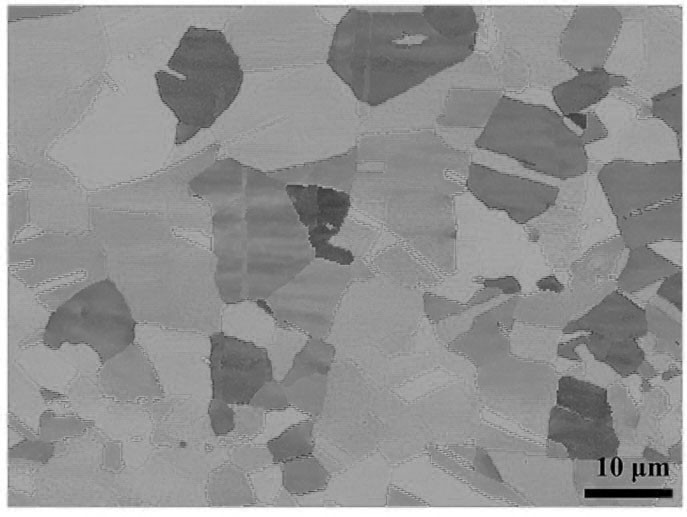}}\hspace{0.1in}
	\subfloat[]{	
	\includegraphics[width=0.2\textwidth]{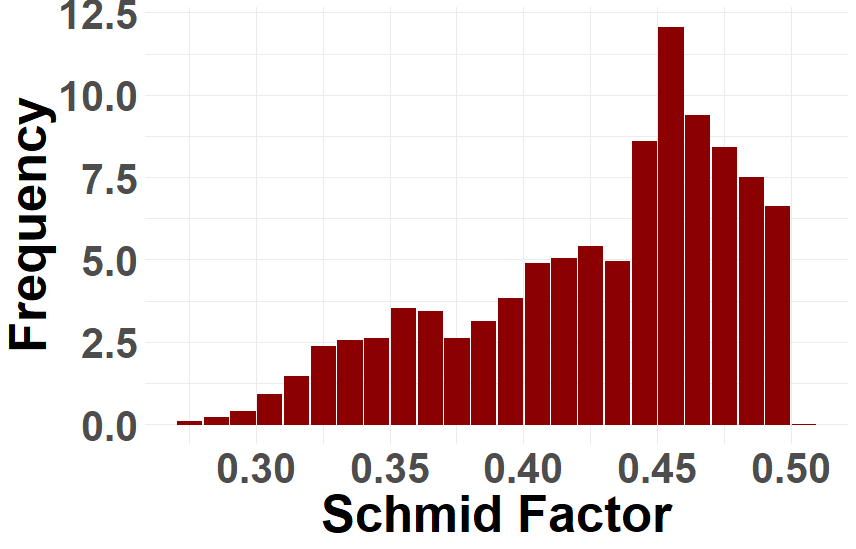}}\hspace{0.1in}
	\includegraphics[width=0.4\textwidth]{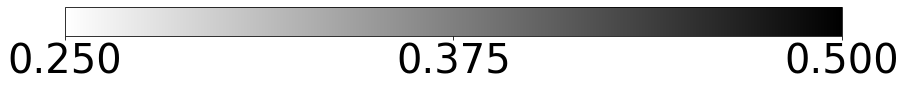}
	
	\caption{Schmid factor of AuSS annealed for 60 seconds at different temperatures.
	(a), (b): 700 $^{\circ} C$.
	(c), (d): 750 $^{\circ} C$.
	(e), (f): 800 $^{\circ} C$.
	(g), (h): 850 $^{\circ} C$.
	(i), (j): 950 $^{\circ} C$.
	The gray scale bar is associated with the range of values of the Schmid factor.}
	\label{fig:app_1}
\end{figure}

The Schmid factor is used to identify grains that are prone to deformation and that may consequently  result in lower material strength \cite{Li2019}. 
% are directly associated with failure mechanisms to  microstructure  \cite{CHEN2013}. Precisely, a Schmid factor of less than 0.35 is categorized as soft grain and deformation occurs easily in soft grains, resulting in lower yield strength \cite{Li2019}. 
On the other hand, 
the annealing temperature in the processing of NG structured materials impacts material strength and the microstructure properties, such as grain size. A quantitative study of the relationship among annealing temperature, the Schmid factor and materials properties can enhance understanding of materials' strength. However, 
one of the main obstacles  is the limited number of data to perform statistical analysis. 
% The authors of \cite {Na2021} lay the foundation towards this research direction through a novel application of persistent homology (PH). However, due to the availability of one set frequency distributions of the Schmid factor, no statistical analysis can be performed.
To overcome this challenge, we apply RPDG to produce a sequence of PDs from the one generated by the
empirical distribution of the Schmid factor (Figure \ref{fig:app_1}).

\subsection{Data} \label{subsect:data}

\textcolor{black}{The Schmid factor and its empirical distribution are obtained from an experiment performed on NG structured AuSS. The strips of such steels are cut and annealed at various temperatures ranging from 700 $^{\circ} C$ to 950 $^{\circ} C$. 
 Figure \ref{fig:app_1} shows the distribution of Schmid factor for varying annealing temperatures.
 % The colors represent values of  the Schmid factor, where
 Light and dark gray regions
 represent lower and higher Schmid factor values, respectively.
 The authors of \cite {Na2021} have shown that
 steel strength and ductility improves
 when the annealing temperature ranges between
 700-850 $^{\circ} C$,
 whereas steel strength decreases
 when the annealing temperature increases to 950 $^{\circ} C$.} Indeed, the goal of this analysis is to reveal a similar behavior at temperatures 700-850 $^{\circ} C$ with the breaking point being at 950 $^{\circ} C$.
%  the fraction of high-angle grain boundaries (HAGBs) is high which prevents the generation of microcracks and  exhibit higher ductility of NG/UFG steels. 
% Quantitative analysis of annealed samples of NG steels enables the development of a predictive relationship between
% processing (annealing temperature), structure (distribution of Schmid factor), and property (strength).}
 
  \begin{table}
		% \vspace{-15pt}
\captionsetup{name=TABLE, justification=centering,labelsep=newline}
	\begin{center}
	\caption{$p-$values based on KS testing without RPDG,
	as described in Section \ref{subsec:vanilla_KS}.  No
statistically significant difference between
associated temperatures is shown at the  threshold 0.05.} \label{tab:hypothesis_test_schmid}
	\begin{tabular}{|c|c|c|c|c|}
			\hline
		& ${750^{\circ}C}$ & ${800^{\circ}C}$ & ${850^{\circ}C}$ & ${950^{\circ}C}$  \\
			\hline
		${700^{\circ}C}$ & 0.068 &  0.259 & 0.441 & 0.139\\
		\hline
		${750^{\circ}C}$ & & 0.992 & 0.893 & 0.675 \\
		\hline
		${800^{\circ}C}$ & & & 0.893 & 0.893\\
		\hline
		${850^{\circ}C}$ & & & & 0.893 \\
			\hline
		\end{tabular}
	\end{center}
	%	\vspace{10pt}
	\end{table}

 \subsection{KS testing without RPDG} \label{subsec:vanilla_KS}
 
We have one histogram (empirical distribution)
of the Schmid factor per annealing temperature,
as shown in Figure \ref{fig:app_1} (right column).
 A two-sided Kolmogorov-Smirnov (KS) hypothesis test is performed
 for each pair of these empirical distributions of Schmid factors,
 and the associated $p$-values are reported in Table \ref{tab:hypothesis_test_schmid}.
 It is noted that all of the $p-$values are higher than the significance level of $0.05$, thus indicating that the materials are showing similar behavior. This agrees with the experimental knowledge \cite{Na2021}
 for the annealing temperatures of 700$^{\circ} C$, 750 $^{\circ} C$, 800 $^{\circ} C$ and 850 $^{\circ} C$, yet their behavior should have changed at the annealing temperature of 950 $^{\circ} C$.
 Thus, the KS hypothesis tests based on the empirical distributions of the Schmid factors for this experiment fail to uncover the important different behavior of materials at 950 $^{\circ} C$ as shown  in \cite{Na2021}.
 
\begin{table}
\captionsetup{name=TABLE, justification=centering,labelsep=newline}
% \vspace{-20pt}
	\begin{center}
	\caption{$p-$values based on KS testing with RPDG,
	as described in Section \ref{subsec:KS_RPDG}.
	Bold indicates statistically significant difference
	between associated temperatures.
	The selected $p$-value threshold is $0.05$.}
% 	It is known from experimental results~\cite{Na2021}
% 	that differences are observed only for the temperature
% 	pairs related to the last column.
% 	Thus, KS testing with RPDG
% 	captures significant differences (bold cells)
% 	only when such differences are present in the experimental results,
% 	aside from the pairing of temperatures at 700$^{\circ}C$ and 800$^{\circ}C$.}
	\label{tab:hypothesis_test_summary}
	\begingroup
{		\begin{tabular}{|c|c|c|c|c|}
        
			\hline
		& ${750^{\circ}C}$ & ${800^{\circ}C}$ & ${850^{\circ}C}$ & ${950^{\circ}C}$  \\
			\hline
			${700^{\circ}C}$ & 0.063 & \textbf{0.002} & 0.123 & \textbf{0.002}\\
		\hline
			${750^{\circ}C}$ & & 0.572 & 0.791 & $\mathbf{5.932\times 10^{-5}}$ \\
		\hline
			${800^{\circ}C}$ & & & 0.123 & $\mathbf{3.471\times 10^{-6}}$\\
		\hline
			${850^{\circ}C}$ & & & & $\mathbf{2.099\times 10^{-4}}$ \\
			\hline
		\end{tabular}}
		\endgroup
	\end{center}
	%	\vspace{10pt}
	\end{table}
	
\subsection{KS testing with RPDG}
\label{subsec:KS_RPDG}

Using the sublevel set filtration of Section \ref{subsec:sl}, 
we generate five persistent diagrams (PDs) from the different empirical distributions of Schmid factors (based on the histograms in Figure \ref{fig:app_1})  corresponding to the five different annealing temperatures.
% , and denote them by
% ${D}^{org}_{{AT}_i},~i=1,\ldots,5$, respectively. 
%  By viewing the frequency distribution as a bounded continuous function of Schmid factor $f(s)$, the sublevel set filtration tracks the evaluation of connected components in sets  $f^{(-1 )} (-\infty,r]$, as $r$, a frequency value, varies from a local minimum to a local maximum. 
For each annealing temperature, we generate a sample of $100,000$ PDs using RPDG with an empirically chosen normal mixture as the proposal density.
We then generate a histogram of persistence values for each set of $100,000$ PD samples,
thereby obtaining five histograms.
Subsequently, we perform a KS test for each pair of histograms
and report the associated $p-$values in Table \ref{tab:hypothesis_test_summary}.
Notice that aside from the pairing at annealing temperatures 700$^{\circ}C$ and 800$^{\circ}C$, all pairings agree with the experimental results \cite{Na2021}, and most importantly, our approach using RPDG reveals the different behavior at  950$^{\circ}C$.
Hence, the hypothesis test based on RPDG sampling
can robustly establish a relationship between processing (annealing temperature),
structure (distribution of the Schmid factor), and property (strength).

\section{Conclusions} 
\label{sec:conclusion}

Data generation in experimental facilities may be expensive,
and thus a small number of noisy data may be collected.
Small sample size poses limitations to statistical analysis.
To remedy this, we have proposed in this work
random persistence diagram generation (RPDG),
a method that randomly generates persistence diagrams (PDs)
by retaining the topological properties encoded
by the PD of a given dataset. {{An interesting theoretical direction is to examine if distributions of PDs generated by RPDG are stable under small perturbations of the initial PD, and in addition study rates of convergence in distribution.}}

RPDG makes two main contributions,
a model of PDs based on a \emph{novel} pairwise interacting point process
and the \emph{first} reversible jump MCMC (RJ-MCMC) technique
for sampling PDs.
It is typical for PDs to have a varying number of points,
and  RJ-MCMC accommodates the randomness
in the location and number of points. 
{{The RPDG method currently treats parameter estimation in its PD model as a pre-processing step, e.g., the jump points in Definition \ref{def:step_int} are empirically selected. A line of future research is to develop a Bayesian version of RPDG,
which will account for uncertainty in its model parameters and will automatically estimate them.}}

Finally, we have employed our RPDG method to
elicit from experimental data~\cite{Na2021}
the relationship of materials' strength,
as expressed by the Schmid factor,
and annealing temperatures.
As a matter of fact, our RPDG method
matches the experimental knowledge,
providing a modelling framework
that enables the identification of changes in materials structure. 

{{
While RPDG has been applied to the aforementioned materials example in this paper, the main methodology is data-agnostic, and as such,
it can find applications in a plethora of problems, including settings with small sample size, and data with imbalanced classes. To that end, synthetic data generation may be needed for statistical inference, and RPDG could be used for sampling synthetic topological summaries of data.
For example, some healthcare applications involve two classes,
a control and a treatment group;
the treatment group may have a small sample size of pertinent images associated with a rare disease, 
and in turn RPDG could be utilized to counter this limitation by generating persistence diagrams that topologically summarize the shape of these images. 
}}

\section*{Acknowledgements}
{{The authors would like to thank the two anonymous reviewers whose comments substantially improved the manuscript.}} The work has been partially supported by the ARO W911NF-21-1-0094 (VM); NSF DMS-2012609 (VM), and ARL Co-operative Agreement \# W911NF-19-2-0328 (VM). The views and conclusions contained in this document are those of the authors and should not be interpreted as representing the official policies, either expressed or {{implied of}} the Army Research Laboratory or the U.S. Government. The U.S. Government is authorized to reproduce and distribute reprints for Government purposes not withstanding any copyright notation herein.


% \newpage
%\vspace{-0.1in}
\begin{appendices}
\renewcommand{\thesubsection}{\Alph{subsection}}

\subsection{Proof of Proposition \ref{prop_acc}}
\label{app_proof}

% Note: in this sample, the section number is hard-coded in. Following
% proper LaTeX conventions, it should properly be coded as a reference:
%In this appendix we prove the following theorem from
%Section~\ref{sec:textree-generalization}:
		%	To prove Proposition \ref{prop_acc} we use a result about simulation 
		%	method for spatial point processes
%given below; for more details, the reader may refer to \cite{Geyer1994}.
%\begin{proof}[Proof of Proposition \ref{prop_acc}]
Let $D^{(l)} = (d_1^{(l)}, \dots, d_{\vert D^{(l)} \vert}^{(l)})$ be the current PD
at the $l$-th RJ-MCMC iteration. 
A candidate PD $D^*$ is generated
by either relocating $D^{(l)}$
or adding a point to $D^{(l)}$
or  removing a point from $D^{(l)}$.
The derivation of the acceptance probability $a(D^{(l)}, D^*)$
for each of these three moves
is considered separately in this proof.
First, we consider the case of relocating $D^{(l)}$
via Metropolis-Hastings sampling.
For each $i=1,\ldots,\vert D^{(l)}\vert,$
a candidate PD
$D^* = (d_1^{(l)}, \dots, 
d_{i-1}^{(l)}, d_i^*, d_{i+1}^{(l)},\dots,    d_{\vert D^{(l)} \vert}^{(l)})$
is generated
by sampling a new location $d_i^*$ from a proposal density $q$.
The Metropolis-Hastings acceptance ratio is
% \vspace{-0.15in}
\begin{align*} \nonumber
&\rho(D^{(l)}, D^*) 
= \frac{f(D^*\vert\widehat{\theta})q({D}^{(l)})}{f(D^{(l)}\vert\widehat{\theta})q({D}^{*})}\\&= \frac{G({D^*\vert\widehat{\theta}})\big[\prod_{j \neq i} 
s (d_j^{(l)})q({d_j}^{(l)})\big] s (d_i^*) q({d_i}^{(l)})
}{G({D^{(l)}\vert\widehat{\theta}})\big[\prod_{j \neq i} 
s (d_j^{(l)})q({d_j}^{(l)})\big] s (d_i^{(l)}) q({d_i}^{*})} \nonumber\\
&= \frac{Z(\widehat{\theta}) g({D^*\vert\widehat{\theta}}) s (d_i^*) q({d_i}^{(l)})
}{Z(\widehat{\theta}) g({D^{(l)}\vert\widehat{\theta}}) s (d_i^{(l)}) q({d_i}^{*})}, \nonumber
\end{align*}
where $G(D\vert\widehat\theta) = g(D\vert\widehat\theta)/Z(\widehat\theta)$.
The acceptance probability for $D^*$ is
\[a(D^{(l)}, D^*) =\min{\{1, \rho(D^{(l)}, D^*)\}}.\]
Hence, it follows that
the probability $a(D^{(l)}, D^*)$ of accepting $D^*$
is given by Equation \eqref{accp_relocation}.

Next, we consider the case of adding a point $d^*$ to $D^{(l)}$,
with $d^*$ being chosen randomly on $\W$.
The number of points of a PD in $\mathbb{W}$ has a Poisson distribution with mean $\lambda(\mathbb{W})$.
%By noting that $|D^*| = |D^{(l)}|+1$ and
Applying Lemma \ref{lemma},
the acceptance ratio $\rho(D^{(l)}, D^*)$
for the candidate PD $D^* = (D^{(l)},  d^*)$ becomes
	\begin{align*} \nonumber
	&\rho(D^{(l)}, D^*) = \frac{f(D^*\vert\widehat{\theta}) 
	\lambda(\W)}{f(D^{(l)}\vert\widehat{\theta}) (\vert D^{(l)}\vert+1)} \\&= \frac{Z(\widehat{\theta}) g({D^*\vert\widehat{\theta}}) 
	s (d^*) \big[\prod_{j =1}^{\vert D^{(l)}\vert} 
	s (d_j^{(l)})\big] \lambda(\W)}{Z(\widehat{\theta}) 
	g({D^{(l)}\vert\widehat{\theta}}) \prod_{j =1}^{\vert D^{(l)}\vert} 
	s (d_j^{(l)}) (\vert D^{(l)}\vert +1)} \\ \nonumber
	&  = \frac{ f(D^{(l)}\vert \widehat{\theta}) \left[\prod_{i=1}^{\vert D^{(i)}\vert} 
	h_{\widehat{\theta}} 
	(d^{(l)}_i,d^*)\right]s (d^*)  \lambda{(\mathbb{W})}}{f(D^{(l)}\vert \widehat{\theta})(\vert D^{(l)}\vert +1)}\\
	 \nonumber
	 &  = \frac{   \left[\prod_{i=1}^{\vert D^{(i)}\vert} 
	h_{\widehat{\theta}} 
	(d^{(l)}_i,d^*)\right]s (d^*) \lambda{(\mathbb{W})}}{(\vert D^{(l)}\vert +1)}.
	\end{align*}
Hence, the probability
\[a(D^{(l)}, D^*) =\min{\{1, \rho(D^{(l)}, D^*)\}}\]
of accepting $D^* = (D^{(l)},  d^*)$
is given by Equation \eqref{eq:accp_add}.

Last, we consider the case of removing a point $d_i^{(l)} \in D^{(l)}$ from $D^{(l)}$,
with $d_i^{(l)}$ being chosen uniformly at random among the points of $D^{(l)}$.
By applying Lemma \ref{lemma},
the acceptance ratio $\rho(D^{(l)}, D^*)$
for the candidate PD
$D^* = D^{(l)} \setminus  d^{(l)}_i $ takes the form
	\begin{align*} \nonumber
	&\rho(D^{(l)}, D^*)   = \frac{f(D^*\vert \widehat{\theta}) (\vert D^{(l)}\vert -1) 
	}{f(D^{(l)}\vert \widehat{\theta}) (\lambda(\W)} \\& = \frac{Z(\widehat{\theta}) g({D^*\vert \widehat{\theta}}) \prod_{j \neq i} 
	s (d_j^{(l)}) (\vert D^{(l)}-1\vert )}{Z(\widehat{\theta}) 
	g({D^{(l)}\vert \widehat{\theta}}) \prod_{j =1}^{\vert D^{(l)}\vert } 
	s (d_j^{(l)}) \lambda(\W)} \\ \nonumber
	&  = \frac{f((D^{(l)} \setminus  d^{(l)}_i)\vert \widehat{\theta})(\vert D^{(l)}-1\vert )} { 
	f(D^{(l)}\vert \widehat{\theta}) \lambda{(\mathbb{W})}}\\ \nonumber
	& =    \frac{\vert D^{(l)}\vert -1} {\left[\prod_{j  \neq i} h_{\widehat{\theta}} 
	(d^{(l)}_j,d_i^{(l)})\right]s (d_i^{(l)}) \lambda{(\mathbb{W})}}.
	\end{align*}
Hence, the probability
\[a(D^{(l)}, D^*) =\min{\{1, \rho(D^{(l)}, D^*)\}}\]
of accepting $D^* = D^{(l)} \setminus  d^{(l)}_i $
is given by Equation \eqref{eq:accp_removal}.

\subsection{RPDG setup for knot example}
\label{app_knot}

Table~\ref{tab:mixture_comp_values}
provides the mixture component means $\mu_i$,
variances $\sigma^2_i$
and weights $w_i$ 
of the mixture used as
proposal density
in asymmetric knot example of
Section~\ref{sec:motivating example}.
The mixture component means
have ben set empirically
to be in the vicinity of
persistence values,
as explained in
Section~\ref{subsec:sampling}.

  \begin{table}
		% \vspace{-15pt}
\captionsetup{name=TABLE, justification=centering,labelsep=newline}
	\begin{center}
	\caption{Mixture component means, variances and weights
	of the mixture used as proposal density
	in Section~\ref{sec:motivating example}.}
	\label{tab:mixture_comp_values}
	\begin{tabular}{|c|c|c|c|}
			\hline
	$i$ & $\mu_i$ & $\sigma_i^2$ & $w_i$ \\ \hline
	1 & (0.28, 0.50) & 0.005 & 0.1 \\ \hline
	2 & (0.35, 0.85) & 0.005 & 0.1 \\ \hline
	3 & (0.37, 1.25) & 0.005 & 0.1 \\ \hline
	4 & (0.44, 1.80) & 0.005 & 0.1 \\ \hline
	5 & (0.32, 0.00) & 0.030  & 0.6 \\ \hline
	\end{tabular}
	\end{center}
	%	\vspace{10pt}
	\end{table}

The jumping points
% for the spatial density
used in
target PIPP density $f(\cdot \mid \theta)$
have been set to
$(r_0,r_1,r_2,r_3)=(0,0.3,0.6,0.9)$.
An estimate $\hat\theta$ of $\theta$
has been computed according to
the procedure outlined in Section~\ref{sec:parameter}.

\end{appendices}
\bibliographystyle{sn-mathphys}
\bibliography{sn-bibliography}
\end{document}